\documentclass{article} 
\usepackage{iclr2027_conference,times}

\usepackage{amsmath,amsfonts,bm}

\def\eqref#1{equation~\ref{#1}}

\def\1{\bm{1}}

\DeclareMathAlphabet{\mathsfit}{\encodingdefault}{\sfdefault}{m}{sl}
\SetMathAlphabet{\mathsfit}{bold}{\encodingdefault}{\sfdefault}{bx}{n}

\usepackage{hyperref}
\usepackage{url}

\usepackage{times}
\usepackage{latexsym}
\usepackage{booktabs}
\usepackage[T1]{fontenc}

\usepackage[utf8]{inputenc}

\IfFileExists{microtype.sty}{\usepackage{microtype}}{}

\IfFileExists{inconsolata.sty}{\usepackage{inconsolata}}{}

\usepackage{graphicx}
\usepackage{amsmath}
\usepackage{amssymb}
\usepackage{multirow}

\title{HART: Hybrid Attention-RWKV Time series Foundation Model }

\author{Qingwen Lin$^{1}$, Boyan Xu$^{1}$, Shifeng Xie$^{4,5}$, Zhifeng Hao$^{1,3}$, Keli Zhang$^{4}$, Ruichu Cai$^ {1,2}$\thanks{~~Corresponding author} \\
  \texttt{qingwen\_lin@foxmail.com}, \texttt{hpakyim@gmail.com}, \texttt{shifeng.xie@telecom-paris.fr}, \\
  \texttt{haozhifeng@stu.edu.cn}, \texttt{zhangkeli1@huawei.com}, \texttt{cairuichu@gmail.com} \\[2mm]
  $^{1}$School of Computer Science, Guangdong University of Technology \\
  $^{2}$Peng Cheng Laboratory \\
  $^{3}$College of Science, Shantou University \\
  $^{4}$Huawei Noah's Ark Lab \\
  $^{5}$LIPADE, Universit\'e Paris Cit\'e, Paris, France
}
\iclrfinalcopy 
\begin{document}

\maketitle
\begin{abstract}
Non-autoregressive multi-step forecasting (NAF) enables time series foundation models to predict an entire future horizon in a single forward pass. Despite its strong sequence modeling capabilities, RWKV-7 exhibits long-horizon degradation in our NAF experiments. In this setup, future patches are represented as missing inputs, requiring the recurrent state to evolve across the prediction horizon without new observations. Our analysis of RWKV-7’s generalized delta rule suggests a mechanism we term Placeholder Idling Decay (PID): repeated state transitions can attenuate historical information when placeholder-driven writes provide insufficient compensation.
Motivated by this analysis, we propose HART, a Hybrid Attention-RWKV Time-series Foundation Model that combines direct access to historical tokens with recurrent temporal modeling by alternating attention and RWKV blocks. Layer-wise analysis reveals reduced state-injection norms in most layers, while transition-matrix norms remain similar between historical and prediction regions, consistent with PID.
Controlled experiments on GIFT-Eval and TIME-Benchmark show that HART improves over pure RWKV at medium and long horizons while maintaining comparable short-horizon performance, and achieves better aggregate performance than a comparably sized pure Transformer. Using 36M parameters, approximately 1/28 of Toto-2.0-1B's, HART achieves comparable long-horizon MASE on GIFT-Eval (0.752 vs. 0.754).
\end{abstract}

\section{Introduction}
\begin{figure*}[t!]
    \centering
    \includegraphics[width=\textwidth]{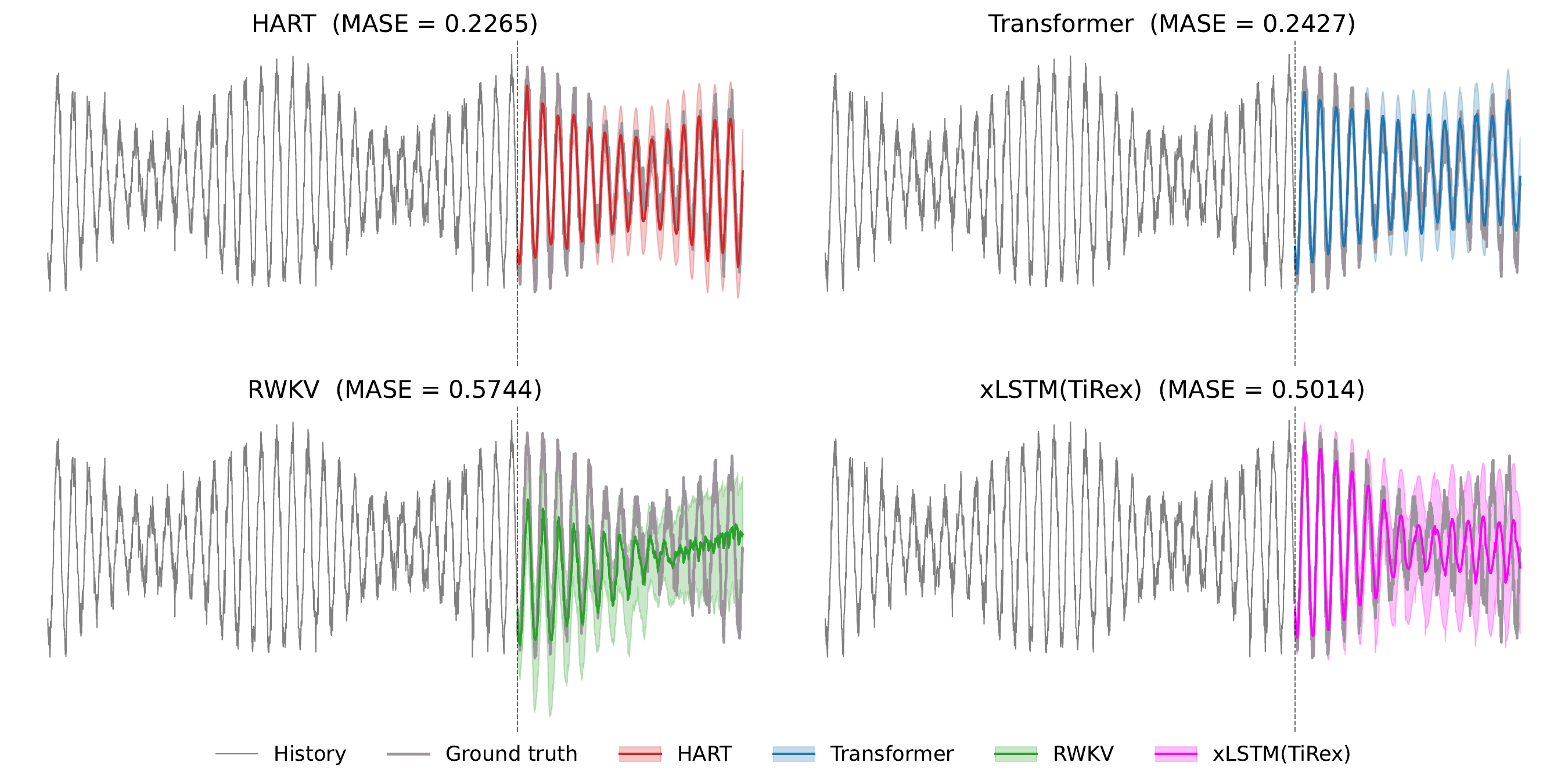}
    \caption{Qualitative NAF predictions on a synthetic signal. Gray: history and ground truth; colored: median predictions with quantile.}
    \label{fig:qualitative-pid}
\end{figure*}
Time series forecasting underpins applications such as energy dispatch, traffic control, and financial risk management~\cite{rizki2026,lee2026fast,wang2026synergizing,liao2026wavelet,wang2026enhancing,kaleem2026lag,xu2024yinglong}.
Non-autoregressive multi-step forecasting (NAF) generates all future steps in a single forward pass, avoiding the error accumulation of autoregressive decoding, and has become the dominant inference paradigm in modern time series foundation models~\cite{auer2026tirex,feng2025kairos,ansari2025chronos,khwaja2026toto}.
NAF, however, imposes one constraint: the prediction region contains no observed input, and the model must rely on its internal state or on retrieval from the historical region.

Recurrent neural networks (RNNs) carry a temporal inductive bias, modeling temporal dependencies through stepwise state updates.
RWKV is one of the strongest pure-RNN language models, matching Transformers of similar scale in language modeling~\cite{peng2023rwkv,peng2025rwkv,peng2024eagle}, with long-range modeling ability validated at scale.
On these grounds, it should be competitive in long-horizon time series forecasting.
However, our NAF experiments show that RWKV-7 exhibits systematic degradation at long horizons.

Figure~\ref{fig:qualitative-pid} shows that RWKV-7 collapses in amplitude over the prediction region on a synthetic signal, while the Transformer shows no comparable collapse under the same data and training configuration. This contrast motivates examining how recurrent state updates interact with placeholder inputs.
In the NAF prediction region, placeholders carry positional information but contain no observed future values. When placeholder-driven writes provide insufficient compensation for repeated state transitions, historical representations in a fixed-capacity recurrent state can be progressively attenuated, and predictions can lose their dynamic structure and drift toward the sequence mean.
We term this mechanism \emph{Placeholder Idling Decay} (PID).

Attention can directly query historical patches without traversing the recurrent state's stepwise decay path.
Based on this distinction, we propose \textsc{HART} (Hybrid Attention-RWKV Time-series Foundation Model): RWKV blocks perform temporal compression and local modeling in the historical region, while Transformer blocks retrieve historical information in the prediction region through global attention, bypassing the stepwise decay path. The two are stacked alternately, providing both local modeling and global access.

Layer-wise analysis shows reduced state-injection norms in most layers, while transition-matrix norms remain similar between historical and prediction regions, consistent with PID.
On GIFT-Eval and TIME-Benchmark, HART significantly improves over pure RWKV at medium and long horizons while maintaining comparable short-horizon performance.
HART also outperforms the controlled pure Transformer baseline on aggregate metrics, supporting the benefit of combining attention with recurrence.
TiRex also shows a similar collapse in the placeholder region in Figure~\ref{fig:qualitative-pid}; HART-36M, with a parameter count and pre-training task aligned with TiRex, achieves better long-horizon and overall metrics than TiRex (GIFT Long: 0.752 vs 0.767, All: 0.710 vs 0.716; TIME Long: 0.7233 vs 0.7608, All: 0.6658 vs 0.6831), with comparable short-horizon performance.

Our main contributions are as follows:
\begin{enumerate}
    \item We characterize Placeholder Idling Decay (PID) in RWKV-7 under NAF through generalized delta-rule analysis and layer-wise norm measurements.
    \item We propose \textsc{HART}, which bypasses the recurrent decay path by alternately stacking RWKV and Transformer blocks.
    \item On GIFT-Eval and TIME-Benchmark, HART achieves better long-horizon and overall MASE than comparably sized baselines.
\end{enumerate}

\section{Related Work}

Time series forecasting is widely applied in energy, transportation, finance, and healthcare, where deep learning has become the mainstream approach~\cite{asghar2024artificial,zhu2025long,suresh2025benchmarking,zhu2023time,lu2025ct,shao2024st,cao2026mstf}.
We review time series foundation models, with an emphasis on recurrent architectures, followed by hybrid recurrent-attention models.

\subsection{Time Series Foundation Models}

Early forecasting was dominated by autoregressive models, such as Lag-Llama~\cite{rasul2023lag}, TimesFM~\cite{das2023decoder}, and YingLong~\cite{xu2024yinglong}, but their error accumulates with the forecast horizon.
NAF generates all future steps in a single forward pass and substantially alleviates this problem, and has been widely adopted by TiRex~\cite{auer2026tirex}, Toto~\cite{khwaja2026toto}, Chronos~\cite{ansari2025chronos}, Xihe~\cite{sun2025xihe}, TempoPFN~\cite{moroshan2025tempopfn}, TabPFN-TS~\cite{hoo2025tables}, Moirai~\cite{liu2025moirai}, Sundial~\cite{pmlr-v267-liu25be}, and Kairos~\cite{feng2025kairos}.

TiRex~\cite{auer2026tirex} is the most closely related baseline: it is built on xLSTM, has a parameter count comparable to our 36M model, and uses the same CPM pre-training task and Pinball loss.
It is among the strongest fixed-capacity recurrent models on NAF benchmarks.
We compare HART with TiRex using results from the official benchmark leaderboards, as reported in Section~\ref{sec:benchmark-comparison}.

RWKV is a pure RNN architecture that matches Transformers of similar scale in language modeling~\cite{peng2023rwkv,peng2024eagle,peng2025rwkv}.
Several works have applied RWKV to time series~\cite{rizki2026,lee2026fast,wang2026synergizing}, but they target specific applications and do not develop it into a time series foundation model; other linear recurrent models such as xLSTM-Mixer~\cite{kraus2026xlstm} show similar potential.
We examine how NAF inference affects recurrent state updates in RWKV-7, particularly in the placeholder-driven prediction region. This question motivates our analysis of PID in Section~\ref{sec:method-pid}.

\subsection{Hybrid Recurrent-Attention Architectures}

Mixing recurrent models with attention is a well-established architecture family in the language domain, including Kimi K3~\cite{team2026kimi} and Jamba~\cite{lenz2025jamba}; ST-MambaSync~\cite{shao2025st} applies a similar idea to time series.

Hybrid architectures have been developed to balance modeling capacity and inference efficiency.
Our work uses the PID analysis of RWKV-7 under NAF to motivate hybrid design: attention bypasses the stepwise decay path in the prediction region, while RWKV preserves the local temporal inductive bias in the historical region. The detailed design is given in Section~\ref{sec:method-arch}.

\section{Method}\label{sec:method-rwkv}

This section first formulates the NAF setting and analyzes PID in RWKV-7, then introduces the \textsc{HART} architecture motivated by this analysis, and finally presents the training objective and inference pipeline.

\subsection{Non-Autoregressive Forecasting and Placeholder Idling Decay}\label{sec:method-pid}

\paragraph{Formalization of Non-Autoregressive Multi-Step Forecasting}

Given a historical sequence $X_{1:C} = (x_1, \ldots, x_{C})$, the goal of NAF is to predict the next $H$ steps $\hat{X}_{C+1:C+H}$ in a single forward pass.
The model input is the concatenation of historical patches and prediction placeholders: for each position in the prediction region, the original scalar input is set to $0$ and the missingness indicator is set to $1$, marking the position as a placeholder to be predicted. The original input, observation mask, and missingness mask are concatenated and passed through the patch embedding layer to obtain the vector representation of that position.
The placeholder embedding incorporates the masks and is not constrained to be zero; the prediction region contains no observed future values.
$P$ denotes the patch length, $T_{hist} = \lceil C/P \rceil$ and $T_{pred} = \lceil H/P \rceil$ the numbers of historical and prediction patches, $D$ the hidden dimension, $h$ the number of heads, and $d_h = D/h$ the per-head dimension.
This placeholder construction is used throughout the paper.
NAF excludes feeding predicted values back into later forecast inputs. A recurrent backbone can still perform successive state updates within the single forecasting pass.

\paragraph{State Update under the Generalized Delta Rule}

RWKV-7~\cite{peng2025rwkv} maintains a recurrent state matrix $S_t \in \mathbb{R}^{d_h \times d_h}$ as a fast-weight memory for key-value associations and updates it via the generalized delta rule (GDR):

\begin{equation}\label{eq:gdr}
S_t = S_{t-1} \underbrace{\bigl( \mathrm{diag}(w_t) - \hat{\kappa}_t^\top (a_t \odot \hat{\kappa}_t) \bigr)}_{M_t} + \underbrace{v_t^\top \tilde{k}_t}_{K_t}
\end{equation}

All vectors in this update are per-head row vectors: $a_t$ is the in-context learning-rate vector, $\hat{\kappa}_t$ is the normalized removal key, and $\tilde{k}_t$ is the replacement key.
The transition matrix $M_t$ propagates and transforms the previous state, while the outer-product term $K_t = v_t^\top \tilde{k}_t$ writes new key-value associations.
These quantities are computed from the layer's token representations through token shift, learned projections, and nonlinear transformations; their parameterization is given in Appendix~\ref{sec:module-details}.

\paragraph{Historical-State Propagation across the Prediction Region}

Let $t_0=T_{hist}$ denote the last historical patch and define the ordered product $\Pi_{i:j}=M_i M_{i+1}\cdots M_j$, with $\Pi_{i:j}=I$ when $i>j$.
Unrolling Equation~\ref{eq:gdr} over $m$ prediction patches gives
\begin{equation}\label{eq:state-unroll}
S_{t_0+m}=S_{t_0}\Pi_{t_0+1:t_0+m}
+\sum_{j=1}^{m}K_{t_0+j}\Pi_{t_0+j+1:t_0+m}.
\end{equation}
The first term propagates the boundary state through all prediction transitions; each later write passes through the remaining transitions. This identity alone implies no contraction, and later writes can also carry history supplied by preceding layers.
Equal transition norms can nevertheless act on different state directions: for example, $\operatorname{diag}(1,0)$ and $\operatorname{diag}(0,1)$ both have unit Frobenius norm but preserve different columns of a right-multiplied state. The product in Equation~\ref{eq:state-unroll} therefore matters beyond individual matrix magnitudes.

\paragraph{Degradation under NAF and PID}

Under NAF, the model receives observed values in the historical region and placeholders in the prediction region.
Although placeholder representations can carry historical context from preceding layers, no new observations enter during prediction.
Equation~\ref{eq:state-unroll} separates this historical-state propagation from subsequent writes.
If these transitions attenuate that contribution and placeholder-driven writes provide insufficient compensation, useful historical information can diminish as $T_{pred}$ grows.
We term this potential failure mode \emph{Placeholder Idling Decay} (PID).
PID concerns insufficient write compensation during recurrent state evolution and does not require accelerated decay in the prediction region.

To examine this hypothesis, we compare the Frobenius norms of $K_t$ and $M_t$ between the historical and prediction regions in Section~\ref{sec:mechanistic-evidence}.
Reduced write norms alongside similar transition norms would be consistent with PID.
These measurements characterize matrix magnitudes; they do not directly establish historical information loss or the contraction rate of transition products.

Our mechanistic analysis is restricted to RWKV-7.
The xLSTM-based TiRex~\cite{auer2026tirex} exhibits qualitatively similar degradation in Figure~\ref{fig:qualitative-pid}, but whether the same mechanism applies to TiRex or other recurrent architectures remains unverified.

\subsection{\textsc{HART}: A Mechanism-Driven Hybrid Architecture}\label{sec:method-arch}

\begin{figure*}[t]
    \centering
    \includegraphics[width=0.80\linewidth]{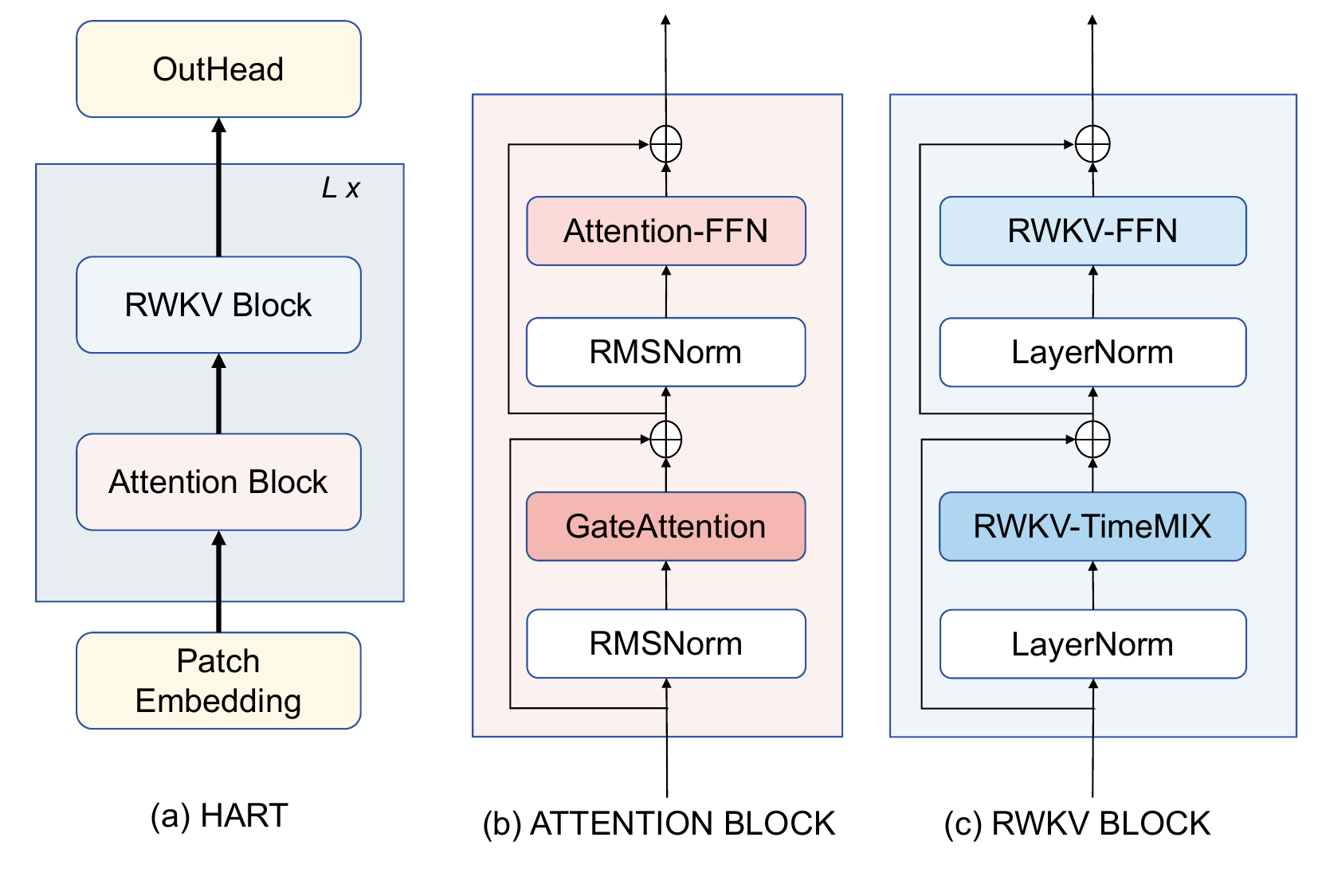}
    \caption{Architecture of the proposed hybrid backbone. (a)~Overall pipeline: time-patch embedding feeds into an alternating stack of Attention and RWKV blocks ($L$ times), followed by the output head. (b)~Attention block: RMSNorm, gated self-attention, and feed-forward with residual connections. (c)~RWKV block: LayerNorm, RWKV-TimeMIX (generalized delta rule), and feed-forward with residual connections.}
    \label{fig:architecture}
\end{figure*}

The PID analysis in Section~\ref{sec:method-pid} motivates direct access to historical tokens during NAF prediction.
Attention provides this access without propagating information through every intervening recurrent state; the Transformer in Figure~\ref{fig:qualitative-pid} also shows no comparable collapse in the illustrated example.
\textsc{HART} combines recurrent temporal modeling with global information retrieval by alternating RWKV and Transformer blocks.
Both block types process the full token sequence: RWKV updates a recurrent representation, while attention allows prediction tokens to retrieve historical information directly.

\paragraph{Overall Architecture}

As shown in Figure~\ref{fig:architecture}, \textsc{HART} consists of three components: a patch embedding layer, an alternately stacked backbone, and an output head.
The backbone contains $L$ alternately stacked module groups, each consisting of one Transformer block followed by one RWKV block, giving $2L$ layers in total.
This order gives each RWKV block attention-conditioned representations at prediction positions. Its writes can therefore depend on retrieved history even when the original future values are masked. Repeating the pair makes this route available at successive depths, supplementing the historical information carried by the recurrent state.
We make only minimal modifications to the Transformer and RWKV blocks; the complete structure is given in Appendix~\ref{sec:module-details}.

\paragraph{Patch Embedding Layer}

The input series is first normalized by $z$-score and then segmented into patches of length $P$.
Each patch is concatenated with the observation mask and the missingness mask, and then projected to dimension $D$ through the skip embedding of Moirai~\cite{liu2025moirai}.
The resulting embeddings serve as tokens for the subsequent Attention and RWKV blocks; the complete formulation is given in Appendix~\ref{sec:patch_details}.

\paragraph{Responsibilities of Transformer and RWKV Blocks}

Transformer blocks perform global information retrieval.
In the prediction region, the attention mechanism can associate placeholder positions with all historical tokens, bypassing the stepwise decay path of the recurrent state and retrieving relevant historical patterns.
These blocks adopt gated self-attention and SwiGLU feed-forward networks; the complete formulation is given in Appendix~\ref{sec:att-block}.

RWKV blocks provide recurrent temporal modeling.
In the historical region, the GDR accumulates and updates observational information in the state matrix.
In the prediction region, the inputs to each RWKV block have been contextualized by the preceding Transformer block, allowing historical information retrieved by attention to enter the recurrent computation.
These blocks adopt RWKV-TimeMIX (GDR) and SwiGLU feed-forward networks; the complete formulation is given in Appendix~\ref{sec:rwkv-block}.

\paragraph{Output Head}

The backbone output is passed through LayerNorm, mapped by a residual MLP to $|Q_{\text{quantile}}|=9$ quantile predictions, and restored to the original scale by inverse $z$-score normalization.

\subsection{Training Objective and Pipeline}\label{sec:method-training}

\paragraph{Pre-training Task}

We adopt contiguous patch masking (CPM) with next-patch prediction, following TiRex~\cite{auer2026tirex}; related NAF pre-training approaches include TiRex, Toto, and TempoPFN~\cite{auer2026tirex,khwaja2026toto,moroshan2025tempopfn}.
During training, contiguous groups of patches are randomly masked, and the model reconstructs the masked regions from the available context. Masked patches are replaced by placeholders in the input, and their ground-truth values serve as supervision in the loss; unmasked positions also contribute through next-patch prediction.
CPM exposes the model to the NAF inference mode during training: it learns to reconstruct a masked future from arbitrary historical positions rather than extrapolating only at the sequence terminus. The detailed construction is given in Appendix~\ref{sec:cpm}.

\paragraph{Quantile Loss}

The model outputs $|Q_{\text{quantile}}|=9$ equidistant quantiles $\{0.1, 0.2, \ldots, 0.9\}$.
Given the ground truth $y_t$ and the quantile predictions $\hat{y}_t^q$, we adopt the Pinball loss:

\begin{equation}\label{eq:pinball}
\mathcal{L} = \sum_{t \in \mathcal{M}} \sum_{q \in Q_{\text{quantile}}}
\begin{cases}
q\,(y_t - \hat{y}_t^q) & \text{if } \hat{y}_t^q \leq y_t \\
(1 - q)\,(\hat{y}_t^q - y_t) & \text{otherwise}
\end{cases}
\end{equation}

where $\mathcal{M}$ denotes the set of positions whose original values exist, so that training combines both the CPM and autoregressive tasks. The loss is computed on the standardized scale, reducing its dependence on the original magnitude of each series.

\paragraph{Inference Pipeline}

At inference, the model switches to the NAF mode: the input is the concatenation of historical tokens and prediction placeholders, and all prediction steps are generated in a single forward pass.
After inference, the mean and standard deviation of the historical sequence are used for inverse $z$-score normalization to restore the output to the original scale.
Training hyperparameters, pre-training data composition, and processing details are given in Appendix~\ref{sec:exp-setup}.

\section{Experiments}
\label{sec:experiments}

\subsection{Evaluation Setup}

All models are evaluated on two benchmarks: GIFT-Eval~\cite{aksu2024gift} and TIME-Benchmark~\cite{wei2026time}.
The GIFT-Eval results cover 97 evaluation entries across Short, Medium, and Long horizon groups; per-dataset results and horizon details are provided in Appendix~\ref{sec:detail-result}.
TIME-Benchmark contains 98 task-frequency combinations (39 unique datasets after deduplication), comprising 50 Short tasks (averaging 30 steps), 24 Medium tasks (averaging 122 steps), and 24 Long tasks (averaging 367 steps).
We adopt MASE (Mean Absolute Scaled Error) as the primary metric, computed relative to the seasonal naïve forecast following the official protocol, and aggregate results using the geometric mean.

We train three models under strictly controlled conditions: HART, pure RWKV, and pure Transformer. All operate in the NAF mode: the input is the concatenation of historical patches and prediction placeholders, and all prediction steps are produced in a single forward pass. Except for the core blocks, all other components remain identical across the three models; detailed training configurations are provided in Appendix~\ref{sec:exp-setup}.
Results for external baselines are obtained from the official leaderboards of GIFT-Eval and TIME-Benchmark.

\subsection{Mechanistic Evidence of PID}\label{sec:mechanistic-evidence}

Following the diagnostic analysis in Section~\ref{sec:method-pid}, we compare write and transition magnitudes across the historical and prediction regions.
Table~\ref{tab:pid-norms} reports the layer-wise mean Frobenius norms of a pure RWKV model, averaged over 4,096 synthetic sequences.
For this diagnostic, the history contains 16,384 time steps and the prediction region contains 1,024, using patches of length 16. Norms are averaged over heads and sequences before the regional comparison; the temporal traces in the appendix show variation within these averages.

\begin{table*}[t]
\centering
\caption{Layer-wise mean Frobenius norms of $K_t$ and $M_t$ in the historical and prediction regions, averaged over 4,096 sequences. Lower ratios indicate smaller prediction-region norms relative to the historical region.}
\label{tab:pid-norms}
\small
\begin{tabular*}{\textwidth}{@{\extracolsep{\fill}}lcccccc@{}}
\toprule
 & \multicolumn{3}{c}{$K_t$} & \multicolumn{3}{c}{$M_t$} \\
\cmidrule(lr){2-4} \cmidrule(lr){5-7}
Layer & Hist. & Pred. & Ratio & Hist. & Pred. & Ratio \\
\midrule
1  & 8.78 & 6.91 & 0.79 & 6.77 & 6.98 & 1.03 \\
2  & 4.72 & 3.10 & 0.66 & 7.20 & 7.29 & 1.01 \\
3  & 4.72 & 2.93 & 0.62 & 7.26 & 7.39 & 1.02 \\
4  & 6.00 & 5.25 & 0.87 & 7.24 & 7.32 & 1.01 \\
5  & 7.15 & 4.29 & 0.60 & 7.17 & 7.27 & 1.01 \\
6  & 5.22 & 4.45 & 0.85 & 7.31 & 7.34 & 1.00 \\
7  & 9.52 & 8.01 & 0.84 & 7.28 & 7.34 & 1.01 \\
8  & 3.51 & 3.40 & 0.97 & 7.36 & 7.44 & 1.01 \\
9  & 6.10 & 5.73 & 0.94 & 7.20 & 7.23 & 1.00 \\
10 & 6.80 & 6.44 & 0.95 & 7.27 & 7.40 & 1.02 \\
11 & 7.76 & 6.55 & 0.84 & 7.26 & 7.45 & 1.03 \\
12 & 3.61 & 4.83 & 1.34 & 7.25 & 7.46 & 1.03 \\
\bottomrule
\end{tabular*}
\end{table*}
The write norms exhibit a clear layer-wise pattern. In the first seven layers, the prediction-to-history ratio of $K_t$ falls between 0.60 and 0.87; in Layers 8 to 11, the ratios range from 0.84 to 0.97. Layer 12 exhibits a reversal, with a ratio of 1.34.
Thus, prediction-region mean write norms are lower in the first eleven layers, with the strongest relative reductions occurring in the earlier layers.

Across all layers, the $M_t$ ratio stays within 1.00 to 1.03, indicating similar transition-matrix norms in the two regions.
Together with the lower $K_t$ norms in most layers, this pattern is consistent with the write-compensation hypothesis of PID.
Similar $M_t$ norms alone do not establish equal decay rates or equal retention of historical information.

One possible explanation for the layer-wise variation is that deeper layers receive representations already transformed by preceding layers, so their responses to placeholders can differ from those of the initial layers.
The Layer 12 reversal shows that the change in write magnitude is not uniform across depth.

Reduced write norms do not imply a complete loss of forecasting capability. RWKV remains comparable to Toto-2.0-22M on Long (GIFT Long: 0.770 vs 0.781; TIME Long: 0.755 vs 0.736), while HART improves over RWKV on both benchmarks as reported in Section~\ref{sec:main-result}.
These performance comparisons complement the layer-wise norm diagnostics.
The full experimental setup and additional visualizations are provided in Appendix~\ref{sec:pid-mechanism}.

\subsection{Main Results}\label{sec:main-result}

Table~\ref{tab:main-result} reports geometric-mean MASE for HART, pure RWKV, and pure Transformer on GIFT-Eval and TIME-Benchmark.
We assess the paired median difference between HART and each baseline using bootstrap confidence intervals.
Cells marked with \dag indicate that the 95\% interval excludes zero.
These intervals describe paired median differences, rather than uncertainty in the reported geometric means.

\begin{table*}[t]
\centering
\caption{Self-developed architectures on GIFT-Eval and TIME-Benchmark (MASE, geometric mean). Lower is better; bold marks the lowest displayed value in each column, including ties. \dag indicates that the paired median difference between HART and the baseline has a 95\% bootstrap confidence interval excluding zero.}
\label{tab:main-result}
\small
\setlength{\tabcolsep}{4pt}
\resizebox{\textwidth}{!}{%
\begin{tabular}{@{}lcccccccc@{}}
\toprule
\multirow{2}{*}{Model} & \multicolumn{4}{c}{GIFT-Eval} & \multicolumn{4}{c}{TIME-Benchmark} \\
\cmidrule(lr){2-5} \cmidrule(lr){6-9}
 & Short & Medium & Long & All & Short & Medium & Long & All \\
\midrule
RWKV-36M        & \textbf{0.689} & 0.749\textsuperscript{\dag} & 0.770\textsuperscript{\dag} & 0.719\textsuperscript{\dag} & 0.620 & 0.727\textsuperscript{\dag} & 0.755\textsuperscript{\dag} & 0.676\textsuperscript{\dag} \\
Transformer-35M & 0.695 & 0.747 & 0.763\textsuperscript{\dag} & 0.721\textsuperscript{\dag} & \textbf{0.618} & 0.725\textsuperscript{\dag} & 0.735 & 0.670\textsuperscript{\dag} \\
HART-36M        & \textbf{0.689} & \textbf{0.727} & \textbf{0.752} & \textbf{0.710} & \textbf{0.618} & \textbf{0.716} & \textbf{0.723} & \textbf{0.666} \\
\bottomrule
\end{tabular}%
}
\end{table*}

\paragraph{HART versus Pure RWKV}

HART improves over pure RWKV on Medium and Long horizons while maintaining comparable reported MASE on Short.
Neither comparison between HART and a baseline on Short is marked as significant; the reported geometric means range from 0.689 to 0.695 on GIFT and from 0.618 to 0.620 on TIME.
On GIFT Medium, HART reduces MASE from 0.749 for pure RWKV to 0.727 (a gain of 0.022); on GIFT Long, it reduces MASE from 0.770 to 0.752 (a gain of 0.018). TIME-Benchmark shows gains of 0.011 on Medium and 0.032 on Long.
The paired median comparisons favor HART on Medium and Long in both benchmarks, as indicated by the bootstrap intervals.

The Medium and Long improvements are consistent with the motivation for providing direct access to history during placeholder-based prediction.
The gain is larger on Medium than Long for GIFT, and larger on Long than Medium for TIME.
The shared training configuration supports a comparison of the architectures under the reported setup, but does not isolate PID as the sole cause of the gains.

\paragraph{HART versus Pure Transformer}

The comparison with the pure Transformer evaluates the benefit of combining attention with recurrent blocks.
On aggregate metrics, HART reduces MASE relative to the pure Transformer on both benchmarks: from 0.721 to 0.710 on GIFT All, and from 0.670 to 0.666 on TIME All; in both cases the 95\% bootstrap confidence intervals for the paired median differences exclude zero, supporting the aggregate comparison in favor of HART under the reported evaluation.

At the horizon level, the HART--Transformer paired median difference is significant on GIFT Long and TIME Medium.
The other horizon-specific Transformer comparisons are not marked as significant.
This variation suggests that the benefit of the hybrid depends on the evaluated tasks; the aggregate results alone do not identify the contribution of a particular signal property.

The synthetic experiments in Appendix~\ref{sec:pid-counterfactual} illustrate this task dependence.
On the multi-component SE4 signal, RWKV outperforms both HART and Transformer under NAF (Long: 1.058, 1.148, and 1.151, respectively).
On SE3, HART and Transformer outperform RWKV (Long: 0.265, 0.306, and 0.581, respectively).
These results support combining the two modeling approaches while showing that HART does not dominate every signal regime.

\paragraph{Sensitivity to Available Future Observations}

The same pre-trained checkpoints, without fine-tuning, also compare NAF with teacher-forced AR supplied with real future inputs.
On Long, RWKV's NAF--AR gaps for SE1, SE2, and SE3 are 0.060, 0.095, and 0.316; HART's corresponding gaps are 0.001, 0.030, and 0.096, and the Transformer's are 0.002, 0.016, and 0.134.
RWKV has the largest gap for each of these signals; SE5 and SE6 have absolute gaps of at most 0.001 at the reported precision.
This signal-dependent sensitivity complements the SE4 comparison of absolute forecasting accuracy. Because teacher forcing changes the available information, this control neither isolates PID nor represents deployable autoregressive forecasting.

\subsection{Benchmark Comparison}\label{sec:benchmark-comparison}

Table~\ref{tab:benchmark} compares HART with mainstream foundation models on GIFT-Eval and TIME-Benchmark.

\begin{table*}[t]
\centering
\caption{External baselines on GIFT-Eval and TIME-Benchmark (MASE, geometric mean). Lower is better; bold marks the lowest displayed value in each column, including ties.}
\label{tab:benchmark}
\small
\setlength{\tabcolsep}{3.5pt}
\resizebox{\textwidth}{!}{%
\begin{tabular}{@{}lcccccccc@{}}
\toprule
\multirow{2}{*}{Model} & \multicolumn{4}{c}{GIFT-Eval} & \multicolumn{4}{c}{TIME-Benchmark} \\
\cmidrule(lr){2-5} \cmidrule(lr){6-9}
 & Short & Medium & Long & All & Short & Medium & Long & All \\
\midrule
YingLong-110M\cite{wang2025output}       & 0.766 & 0.856 & 0.885 & 0.809 & -- & -- & -- & -- \\
YingLong-300M\cite{wang2025output}       & 0.758 & 0.841 & 0.867 & 0.798 & -- & -- & -- & -- \\
TimesFM-2.0-500M\cite{das2023decoder}    & 0.704 & 0.810 & 0.860 & 0.758 & 0.652 & 0.791 & 0.797 & 0.718 \\
Sundial-128M\cite{pmlr-v267-liu25be}     & 0.732 & 0.764 & 0.785 & 0.750 & 0.700 & 0.813 & 0.831 & 0.757 \\
Xihe-1.5B\cite{sun2025xihe}              & 0.689 & \textbf{0.718} & 0.763 & 0.711 & -- & -- & -- & -- \\
TempoPFN-40M\cite{moroshan2025tempopfn}  & 0.727 & 0.840 & 0.912 & 0.788 & -- & -- & -- & -- \\
Kairos-23M\cite{feng2025kairos}          & 0.723 & 0.768 & 0.797 & 0.748 & -- & -- & -- & -- \\
Kairos-50M\cite{feng2025kairos}          & 0.719 & 0.759 & 0.789 & 0.742 & 0.683 & 0.804 & 0.802 & 0.740 \\
Moirai2-12.8M\cite{liu2025moirai}        & 0.695 & 0.759 & 0.789 & 0.728 & 0.635 & 0.770 & 0.794 & 0.703 \\
TS-ICL\cite{naour2026ts}                 & -- & -- & -- & -- & 0.639 & 0.749 & 0.765 & 0.694 \\
TiRex-35M\cite{auer2026tirex}            & 0.685 & 0.750 & 0.767 & 0.716 & 0.617 & 0.759 & 0.761 & 0.683 \\
Toto-2.0-22M\cite{khwaja2026toto}        & 0.687 & 0.746 & 0.781 & 0.719 & 0.614 & 0.730 & 0.736 & 0.670 \\
Toto-2.0-313M\cite{khwaja2026toto}       & \textbf{0.671} & 0.736 & 0.759 & 0.703 & \textbf{0.593} & \textbf{0.696} & \textbf{0.706} & \textbf{0.644} \\
Toto-2.0-1B\cite{khwaja2026toto}         & \textbf{0.671} & 0.724 & 0.754 & \textbf{0.699} & 0.595 & 0.700 & \textbf{0.706} & 0.645 \\
\midrule
HART-36M (ours)                          & 0.689 & 0.727 & \textbf{0.752} & 0.710 & 0.618 & 0.716 & 0.723 & 0.666 \\
\bottomrule
\end{tabular}%
}
\end{table*}

TiRex is a closely related recurrent baseline: it has a parameter count comparable to HART (35M vs 36M) and uses CPM pre-training and Pinball loss.
HART supplements recurrence with attention, whereas TiRex uses xLSTM.
Figure~\ref{fig:qualitative-pid} shows qualitatively similar long-horizon degradation in TiRex, without establishing a shared mechanism.
Among TiRex, RWKV, and HART, TiRex has the lowest Short MASE (GIFT Short: 0.685 vs RWKV 0.689 and HART 0.689; TIME Short: 0.617 vs 0.620 and 0.618).
Its difference between the reported Short and Long MASE is larger than HART's (GIFT: 0.082 vs 0.063; TIME: 0.144 vs 0.106).
On Long, HART outperforms both recurrent baselines (GIFT Long: TiRex 0.767, HART 0.752, and RWKV 0.770; TIME Long: 0.761, 0.723, and 0.755, respectively).
These leaderboard comparisons establish the observed ranking across horizon groups, but do not isolate the effects of architecture, training data, or inference design.

In the cross-scale comparison, HART-36M uses approximately 1/28 of Toto-2.0-1B's parameters while performing on par on GIFT Long (0.752 vs 0.754).
Xihe-1.5B uses direct multi-horizon prediction heads and outperforms HART on GIFT Medium (0.718 vs 0.727), while HART achieves lower MASE on Long (0.752 vs 0.763). These results show competitive long-horizon accuracy at a smaller parameter count under the reported benchmark evaluation.

\section{Conclusion}

We investigate RWKV-7 under non-autoregressive multi-step forecasting, where the recurrent state evolves through a prediction region containing placeholders rather than new observations.
Our generalized delta-rule analysis motivates Placeholder Idling Decay (PID): historical information can be attenuated when repeated state transitions are not sufficiently compensated by placeholder-driven writes.
Layer-wise measurements show lower write norms in most layers and similar transition norms across the historical and prediction regions, consistent with this hypothesis.

Motivated by this analysis, \textsc{HART} alternates Transformer and RWKV blocks to combine direct access to historical tokens with recurrent temporal modeling.
Controlled experiments on GIFT-Eval and TIME-Benchmark show improvements over pure RWKV on Medium and Long horizons, comparable Short performance, and lower aggregate MASE than a comparably sized pure Transformer.
Our mechanistic analysis is restricted to RWKV-7; applying the transition--injection decomposition to other recurrent architectures requires separate validation.

\section{Limitations and Future Work}

We analyze the PID hypothesis in RWKV-7 only. The norm diagnostics do not directly measure retained historical information, and the forecasting comparisons do not isolate PID from all other architectural effects.
Figure~\ref{fig:qualitative-pid} shows that the xLSTM-based TiRex also exhibits similar long-horizon amplitude degradation, but whether it shares the same write-compensation mechanism requires per-architecture analysis.
We adopt a 1:1 stacking configuration; whether other ratios or orderings yield further gains remains open.

\section{AI Use Statement}

We used DeepSeek-V4 Pro to assist with writing, polishing, and translation of this manuscript. We also used OpenAI Codex for rhetorical editing, consistency checks, and revision suggestions. The authors take full responsibility for the content and final version.

\newpage
\bibliography{iclr2027_conference}
\bibliographystyle{iclr2027_conference}

\appendix

\section{Detailed Per-Dataset Results}\label{sec:detail-result}

This appendix provides per-dataset MASE scores on GIFT-Eval and TIME-Benchmark for all models reported in the main benchmark table.
Lower MASE indicates better accuracy.
All values are rounded to three decimal places.
Results for external baselines are taken from the official leaderboards of the two benchmarks; HART, RWKV, and Transformer are trained and evaluated by us under identical settings.

\begin{table*}[htbp]
\centering
\caption{Per-dataset MASE on GIFT-Eval long-term datasets (21 datasets).}
\label{tab:long-detail}
\tiny
\setlength{\tabcolsep}{1.5pt}
\resizebox{\textwidth}{!}{%
\begin{tabular}{@{}l*{21}{c}@{}}
\toprule
Model & 
\rotatebox{75}{bitbrains\_fast\_storage/5T} & 
\rotatebox{75}{bitbrains\_rnd/5T} & 
\rotatebox{75}{bizitobs\_application/10S} & 
\rotatebox{75}{bizitobs\_l2c/5T} & 
\rotatebox{75}{bizitobs\_l2c/H} & 
\rotatebox{75}{bizitobs\_service/10S} & 
\rotatebox{75}{electricity/15T} & 
\rotatebox{75}{electricity/H} & 
\rotatebox{75}{ett1/15T} & 
\rotatebox{75}{ett1/H} & 
\rotatebox{75}{ett2/15T} & 
\rotatebox{75}{ett2/H} & 
\rotatebox{75}{jena\_weather/10T} & 
\rotatebox{75}{jena\_weather/H} & 
\rotatebox{75}{kdd\_cup\_2018/H} & 
\rotatebox{75}{loop\_seattle/5T} & 
\rotatebox{75}{loop\_seattle/H} & 
\rotatebox{75}{m\_dense/H} & 
\rotatebox{75}{solar/10T} & 
\rotatebox{75}{solar/H} & 
\rotatebox{75}{sz\_taxi/15T} \\
\midrule
HART-36M          & 0.784 & 0.947 & 0.969 & 0.531 & 0.394 & 0.982 & 0.742 & 0.741 & 0.853 & 1.049 & 0.889 & 0.970 & 1.025 & 0.618 & 0.709 & 0.604 & 0.562 & 0.451 & 0.909 & 0.814 & 0.784 \\
RWKV-36M          & 0.797 & 0.951 & 1.031 & 0.471 & 0.420 & 0.959 & 0.785 & 0.766 & 0.933 & 1.062 & 0.945 & 0.937 & 1.011 & 0.619 & 0.714 & 0.691 & 0.582 & 0.486 & 0.924 & 0.834 & 0.785 \\
Transformer-35M   & 0.783 & 0.947 & 0.977 & 0.636 & 0.397 & 1.090 & 0.755 & 0.744 & 0.847 & 1.029 & 0.902 & 0.883 & 0.979 & 0.626 & 0.736 & 0.621 & 0.567 & 0.454 & 0.935 & 0.820 & 0.784 \\
Kairos-23M        & 0.859 & 0.976 & 0.947 & 0.764 & 0.390 & 1.015 & 0.818 & 0.845 & 0.849 & 0.903 & 0.896 & 0.894 & 0.863 & 0.842 & 0.765 & 0.701 & 0.611 & 0.568 & 0.926 & 0.857 & 0.768 \\
TiRex-35M         & 0.805 & 0.957 & 1.059 & 0.417 & 0.410 & 1.093 & 0.801 & 0.810 & 0.857 & 0.893 & 0.884 & 1.050 & 0.852 & 0.777 & 0.756 & 0.761 & 0.566 & 0.470 & 0.954 & 0.771 & 0.737 \\
Toto-2.0-22M      & 0.736 & 0.934 & 1.037 & 0.693 & 0.402 & 0.967 & 0.787 & 0.821 & 0.838 & 0.914 & 0.875 & 0.925 & 0.852 & 0.837 & 0.762 & 0.714 & 0.583 & 0.523 & 0.954 & 0.859 & 0.741 \\
Toto-2.0-313M     & 0.720 & 0.925 & 1.035 & 0.607 & 0.397 & 0.948 & 0.753 & 0.798 & 0.842 & 0.919 & 0.845 & 0.920 & 0.833 & 0.800 & 0.767 & 0.661 & 0.567 & 0.509 & 0.899 & 0.829 & 0.732 \\
Toto-2.0-1B       & 0.711 & 0.935 & 1.037 & 0.579 & 0.388 & 0.942 & 0.748 & 0.794 & 0.838 & 0.914 & 0.851 & 0.926 & 0.824 & 0.858 & 0.773 & 0.673 & 0.552 & 0.479 & 0.868 & 0.809 & 0.736 \\
TempoPFN-40M      & 0.844 & 0.998 & 1.246 & 0.905 & 0.450 & 1.445 & 1.045 & 0.887 & 1.018 & 0.968 & 1.001 & 0.896 & 0.872 & 0.900 & 0.761 & 0.900 & 0.722 & 0.758 & 1.395 & 0.911 & 0.766 \\
Kairos-50M        & 0.855 & 0.966 & 0.907 & 0.842 & 0.390 & 1.028 & 0.805 & 0.819 & 0.850 & 0.867 & 0.920 & 0.906 & 0.847 & 0.817 & 0.764 & 0.699 & 0.587 & 0.543 & 0.897 & 0.852 & 0.752 \\
Sundial-128M      & 0.890 & 1.006 & 1.156 & 0.437 & 0.466 & 1.066 & 0.779 & 0.706 & 0.914 & 0.950 & 0.910 & 1.010 & 0.892 & 0.858 & 0.580 & 0.714 & 0.639 & 0.522 & 1.090 & 0.700 & 0.778 \\
YingLong-110M     & 0.876 & 0.989 & 1.365 & 0.864 & 0.544 & 1.385 & 0.783 & 0.948 & 0.851 & 0.955 & 0.895 & 0.973 & 0.850 & 0.797 & 0.754 & 0.864 & 0.634 & 0.695 & 1.421 & 0.931 & 0.735 \\
YingLong-300M     & 0.888 & 0.991 & 1.435 & 0.835 & 0.609 & 1.616 & 0.789 & 0.847 & 0.868 & 0.927 & 0.894 & 0.936 & 0.839 & 0.812 & 0.752 & 0.855 & 0.634 & 0.570 & 1.018 & 0.897 & 0.740 \\
Moirai2-12.8M     & 0.807 & 0.952 & 1.069 & 0.399 & 0.428 & 1.003 & 0.780 & 0.828 & 0.896 & 0.991 & 0.947 & 0.932 & 0.963 & 0.820 & 0.801 & 0.742 & 0.585 & 0.486 & 1.161 & 0.825 & 0.768 \\
Xihe-1.5B         & 0.794 & 0.955 & 1.019 & 0.741 & 0.426 & 1.071 & 0.772 & 0.354 & 0.884 & 0.931 & 0.956 & 0.996 & 0.824 & 0.734 & 0.785 & 0.691 & 0.582 & 0.487 & 0.991 & 0.899 & 0.740 \\
TimesFM-2.0-500M  & 0.862 & 1.028 & 1.271 & 0.850 & 0.911 & 1.576 & 0.777 & 0.692 & 0.936 & 1.024 & 0.929 & 0.998 & 0.304 & 0.834 & 0.772 & 0.905 & 0.586 & 0.538 & 1.319 & 1.181 & 0.743 \\
\bottomrule
\end{tabular}%
}
\end{table*}

\begin{table*}[htbp]
\centering
\caption{Per-dataset MASE on GIFT-Eval medium-term datasets (21 datasets).}
\label{tab:medium-detail}
\tiny
\setlength{\tabcolsep}{1.5pt}
\resizebox{\textwidth}{!}{%
\begin{tabular}{@{}l*{21}{c}@{}}
\toprule
Model & 
\rotatebox{75}{bitbrains\_fast\_storage/5T} & 
\rotatebox{75}{bitbrains\_rnd/5T} & 
\rotatebox{75}{bizitobs\_application/10S} & 
\rotatebox{75}{bizitobs\_l2c/5T} & 
\rotatebox{75}{bizitobs\_l2c/H} & 
\rotatebox{75}{bizitobs\_service/10S} & 
\rotatebox{75}{electricity/15T} & 
\rotatebox{75}{electricity/H} & 
\rotatebox{75}{ett1/15T} & 
\rotatebox{75}{ett1/H} & 
\rotatebox{75}{ett2/15T} & 
\rotatebox{75}{ett2/H} & 
\rotatebox{75}{jena\_weather/10T} & 
\rotatebox{75}{jena\_weather/H} & 
\rotatebox{75}{kdd\_cup\_2018/H} & 
\rotatebox{75}{loop\_seattle/5T} & 
\rotatebox{75}{loop\_seattle/H} & 
\rotatebox{75}{m\_dense/H} & 
\rotatebox{75}{solar/10T} & 
\rotatebox{75}{solar/H} & 
\rotatebox{75}{sz\_taxi/15T} \\
\midrule
HART-36M          & 0.805 & 0.965 & 0.765 & 0.463 & 0.331 & 0.817 & 0.717 & 0.745 & 0.845 & 0.889 & 0.830 & 0.898 & 0.932 & 0.840 & 0.691 & 0.619 & 0.620 & 0.424 & 0.903 & 0.890 & 0.774 \\
RWKV-36M          & 0.815 & 0.968 & 0.906 & 0.484 & 0.340 & 0.867 & 0.737 & 0.756 & 0.895 & 0.879 & 0.883 & 0.857 & 0.947 & 0.854 & 0.690 & 0.700 & 0.632 & 0.452 & 0.880 & 0.902 & 0.773 \\
Transformer-35M   & 0.804 & 0.967 & 0.962 & 0.595 & 0.341 & 0.931 & 0.724 & 0.753 & 0.834 & 0.904 & 0.847 & 0.801 & 0.919 & 0.861 & 0.696 & 0.629 & 0.613 & 0.434 & 0.870 & 0.903 & 0.771 \\
Kairos-23M        & 0.886 & 0.989 & 0.936 & 0.605 & 0.325 & 0.906 & 0.788 & 0.818 & 0.854 & 0.840 & 0.865 & 0.814 & 0.882 & 0.944 & 0.726 & 0.702 & 0.661 & 0.511 & 0.812 & 0.900 & 0.781 \\
TiRex-35M         & 0.822 & 0.972 & 0.982 & 0.498 & 0.344 & 0.889 & 0.771 & 0.782 & 0.854 & 0.788 & 0.833 & 0.916 & 0.846 & 0.853 & 0.725 & 0.763 & 0.620 & 0.445 & 0.924 & 0.838 & 0.753 \\
Toto-2.0-22M      & 0.763 & 0.955 & 0.919 & 0.512 & 0.331 & 0.830 & 0.764 & 0.799 & 0.837 & 0.807 & 0.842 & 0.842 & 0.792 & 0.942 & 0.733 & 0.722 & 0.634 & 0.470 & 0.938 & 0.932 & 0.756 \\
Toto-2.0-313M     & 0.752 & 0.949 & 0.926 & 0.543 & 0.322 & 0.802 & 0.719 & 0.766 & 0.832 & 0.797 & 0.826 & 0.854 & 0.795 & 0.920 & 0.734 & 0.658 & 0.616 & 0.462 & 0.957 & 0.937 & 0.748 \\
Toto-2.0-1B       & 0.740 & 0.949 & 0.918 & 0.488 & 0.326 & 0.780 & 0.712 & 0.763 & 0.830 & 0.806 & 0.830 & 0.848 & 0.801 & 0.897 & 0.741 & 0.658 & 0.596 & 0.442 & 0.896 & 0.899 & 0.753 \\
TempoPFN-40M      & 0.892 & 0.998 & 0.879 & 0.692 & 0.363 & 1.131 & 0.928 & 0.853 & 0.993 & 0.829 & 0.902 & 0.857 & 0.857 & 0.988 & 0.751 & 0.948 & 0.672 & 0.635 & 1.107 & 1.024 & 0.778 \\
Kairos-50M        & 0.863 & 0.981 & 0.845 & 0.664 & 0.314 & 0.839 & 0.767 & 0.788 & 0.861 & 0.814 & 0.884 & 0.845 & 0.877 & 0.965 & 0.729 & 0.703 & 0.634 & 0.492 & 0.808 & 0.912 & 0.766 \\
Sundial-128M      & 0.908 & 1.004 & 1.061 & 0.426 & 0.364 & 0.963 & 0.742 & 0.714 & 0.898 & 0.821 & 0.863 & 0.900 & 0.892 & 0.977 & 0.588 & 0.711 & 0.685 & 0.483 & 1.015 & 0.820 & 0.789 \\
YingLong-110M     & 0.869 & 0.994 & 1.362 & 0.725 & 0.434 & 1.384 & 0.765 & 0.934 & 0.850 & 0.809 & 0.851 & 0.831 & 0.895 & 0.996 & 0.724 & 0.896 & 0.659 & 0.605 & 1.228 & 1.005 & 0.753 \\
Moirai2-12.8M     & 0.813 & 0.969 & 0.921 & 0.431 & 0.337 & 0.857 & 0.731 & 0.779 & 0.874 & 0.832 & 0.875 & 0.876 & 0.909 & 0.920 & 0.787 & 0.750 & 0.635 & 0.448 & 1.078 & 0.907 & 0.774 \\
Xihe-1.5B         & 0.815 & 0.970 & 0.624 & 0.657 & 0.349 & 0.772 & 0.748 & 0.362 & 0.876 & 0.793 & 0.888 & 0.863 & 0.866 & 0.910 & 0.752 & 0.695 & 0.613 & 0.450 & 0.948 & 0.918 & 0.753 \\
TimesFM-2.0-500M  & 0.881 & 1.012 & 1.145 & 0.823 & 0.784 & 1.159 & 0.735 & 0.667 & 0.911 & 0.835 & 0.893 & 0.848 & 0.266 & 0.972 & 0.718 & 0.973 & 0.631 & 0.491 & 1.261 & 1.025 & 0.765 \\
\bottomrule
\end{tabular}%
}
\end{table*}

\begin{table*}[htbp]
\centering
\caption{Per-dataset MASE on GIFT-Eval short-term datasets (Part 1: A to M, 28 datasets).}
\label{tab:short-detail-1}
\tiny
\setlength{\tabcolsep}{1.5pt}
\resizebox{\textwidth}{!}{%
\begin{tabular}{@{}l*{28}{c}@{}}
\toprule
Model & 
\rotatebox{75}{bizitobs\_application} & 
\rotatebox{75}{bizitobs\_l2c/5T} & 
\rotatebox{75}{bizitobs\_l2c/H} & 
\rotatebox{75}{bizitobs\_service} & 
\rotatebox{75}{car\_parts/M} & 
\rotatebox{75}{covid\_deaths/D} & 
\rotatebox{75}{electricity/15T} & 
\rotatebox{75}{electricity/D} & 
\rotatebox{75}{electricity/H} & 
\rotatebox{75}{electricity/W} & 
\rotatebox{75}{ett1/15T} & 
\rotatebox{75}{ett1/D} & 
\rotatebox{75}{ett1/H} & 
\rotatebox{75}{ett1/W} & 
\rotatebox{75}{ett2/15T} & 
\rotatebox{75}{ett2/D} & 
\rotatebox{75}{ett2/H} & 
\rotatebox{75}{ett2/W} & 
\rotatebox{75}{hierarchical\_sales/D} & 
\rotatebox{75}{hierarchical\_sales/W} & 
\rotatebox{75}{hospital/M} & 
\rotatebox{75}{jena\_weather/10T} & 
\rotatebox{75}{jena\_weather/D} & 
\rotatebox{75}{jena\_weather/H} & 
\rotatebox{75}{kdd\_cup\_2018/D} & 
\rotatebox{75}{kdd\_cup\_2018/H} & 
\rotatebox{75}{loop\_seattle/5T} & 
\rotatebox{75}{loop\_seattle/D} \\
\midrule
HART-36M          & 0.372 & 0.265 & 0.366 & 0.585 & 0.711 & 0.960 & 0.584 & 0.714 & 0.661 & 0.694 & 0.707 & 0.976 & 0.846 & 0.938 & 0.675 & 0.940 & 0.772 & 0.976 & 0.664 & 0.700 & 0.908 & 0.356 & 0.649 & 0.700 & 0.790 & 0.638 & 0.728 & 0.522 \\
RWKV-36M          & 0.422 & 0.238 & 0.365 & 0.627 & 0.710 & 0.955 & 0.591 & 0.717 & 0.678 & 0.691 & 0.728 & 0.971 & 0.862 & 0.909 & 0.676 & 0.918 & 0.776 & 0.927 & 0.660 & 0.702 & 0.907 & 0.367 & 0.620 & 0.706 & 0.793 & 0.638 & 0.761 & 0.518 \\
Transformer-35M   & 0.488 & 0.270 & 0.361 & 0.635 & 0.705 & 0.972 & 0.569 & 0.713 & 0.668 & 0.704 & 0.710 & 0.972 & 0.853 & 0.938 & 0.669 & 0.898 & 0.774 & 0.978 & 0.657 & 0.705 & 0.907 & 0.373 & 0.638 & 0.707 & 0.784 & 0.627 & 0.749 & 0.525 \\
Kairos-23M        & 0.573 & 0.288 & 0.396 & 0.685 & 0.819 & 0.874 & 0.649 & 0.730 & 0.732 & 0.678 & 0.761 & 0.921 & 0.873 & 0.806 & 0.749 & 0.935 & 0.809 & 1.252 & 0.663 & 0.721 & 0.846 & 0.387 & 0.758 & 0.733 & 0.791 & 0.706 & 0.741 & 0.508 \\
TiRex-35M         & 0.537 & 0.283 & 0.407 & 0.674 & 0.705 & 0.727 & 0.617 & 0.710 & 0.670 & 0.690 & 0.759 & 0.944 & 0.836 & 0.904 & 0.706 & 0.943 & 0.784 & 0.983 & 0.657 & 0.704 & 0.821 & 0.377 & 0.645 & 0.724 & 0.802 & 0.701 & 0.735 & 0.505 \\
Toto-2.0-22M      & 0.542 & 0.254 & 0.340 & 0.631 & 0.735 & 0.632 & 0.596 & 0.737 & 0.779 & 0.773 & 0.713 & 0.930 & 0.871 & 0.857 & 0.669 & 0.926 & 0.802 & 1.054 & 0.661 & 0.711 & 0.823 & 0.351 & 0.711 & 0.729 & 0.801 & 0.706 & 0.730 & 0.514 \\
Toto-2.0-313M     & 0.499 & 0.223 & 0.351 & 0.583 & 0.729 & 0.798 & 0.575 & 0.719 & 0.742 & 0.775 & 0.723 & 0.930 & 0.835 & 0.869 & 0.651 & 0.921 & 0.796 & 0.976 & 0.660 & 0.706 & 0.818 & 0.335 & 0.750 & 0.725 & 0.796 & 0.700 & 0.709 & 0.510 \\
Toto-2.0-1B       & 0.482 & 0.210 & 0.359 & 0.562 & 0.738 & 0.722 & 0.575 & 0.718 & 0.744 & 0.762 & 0.716 & 0.945 & 0.817 & 0.900 & 0.654 & 0.903 & 0.797 & 1.096 & 0.658 & 0.707 & 0.819 & 0.329 & 0.727 & 0.717 & 0.799 & 0.701 & 0.714 & 0.508 \\
TempoPFN-40M      & 0.422 & 0.273 & 0.411 & 0.948 & 0.696 & 0.768 & 0.695 & 0.758 & 0.795 & 0.789 & 0.794 & 0.932 & 0.860 & 0.852 & 0.722 & 1.115 & 0.806 & 1.017 & 0.662 & 0.711 & 0.829 & 0.394 & 0.818 & 0.776 & 0.811 & 0.760 & 0.838 & 0.510 \\
Kairos-50M        & 0.687 & 0.278 & 0.395 & 0.742 & 0.832 & 0.895 & 0.622 & 0.723 & 0.699 & 0.684 & 0.770 & 1.004 & 0.857 & 0.840 & 0.720 & 0.964 & 0.813 & 1.496 & 0.656 & 0.712 & 0.841 & 0.384 & 0.713 & 0.775 & 0.786 & 0.699 & 0.720 & 0.504 \\
Sundial-128M      & 0.637 & 0.252 & 0.392 & 0.685 & 0.796 & 1.287 & 0.521 & 0.733 & 0.686 & 0.773 & 0.760 & 1.070 & 0.849 & 1.042 & 0.700 & 1.084 & 0.835 & 1.203 & 0.696 & 0.733 & 0.910 & 0.400 & 0.592 & 0.745 & 0.784 & 0.597 & 0.711 & 0.519 \\
YingLong-110M     & 0.718 & 0.292 & 0.448 & 0.877 & 0.868 & 0.986 & 0.611 & 0.714 & 0.839 & 0.809 & 0.754 & 0.972 & 0.845 & 0.862 & 0.699 & 0.910 & 0.805 & 1.523 & 0.666 & 0.724 & 0.848 & 0.437 & 0.712 & 0.745 & 0.785 & 0.684 & 0.794 & 0.537 \\
Moirai2-12.8M     & 0.642 & 0.295 & 0.414 & 0.713 & 0.688 & 0.788 & 0.507 & 0.715 & 0.653 & 0.794 & 0.736 & 0.983 & 0.856 & 0.842 & 0.718 & 0.937 & 0.797 & 1.037 & 0.659 & 0.720 & 0.831 & 0.425 & 0.645 & 0.742 & 0.829 & 0.751 & 0.711 & 0.516 \\
Xihe-1.5B         & 0.452 & 0.280 & 0.439 & 0.651 & 0.713 & 0.760 & 0.617 & 0.710 & 0.388 & 0.591 & 0.741 & 0.993 & 0.845 & 0.829 & 0.754 & 0.976 & 0.795 & 1.247 & 0.658 & 0.703 & 0.848 & 0.387 & 0.645 & 0.715 & 0.818 & 0.706 & 0.733 & 0.472 \\
TimesFM-2.0-500M  & 0.697 & 0.317 & 0.644 & 0.846 & 0.767 & 1.010 & 0.528 & 0.748 & 0.562 & 0.693 & 0.769 & 0.928 & 0.886 & 0.933 & 0.700 & 1.122 & 0.818 & 1.443 & 0.663 & 0.686 & 0.820 & 0.122 & 0.790 & 0.727 & 0.809 & 0.702 & 0.765 & 0.496 \\
\bottomrule
\end{tabular}%
}
\end{table*}

\begin{table*}[htbp]
\centering
\caption{Per-dataset MASE on GIFT-Eval short-term datasets (Part 2: M--Z, 27 datasets).}
\label{tab:short-detail-2}
\tiny
\setlength{\tabcolsep}{1.5pt}
\resizebox{\textwidth}{!}{%
\begin{tabular}{@{}l*{27}{c}@{}}
\toprule
Model & 
\rotatebox{75}{loop\_seattle/H} & 
\rotatebox{75}{m4\_daily/D} & 
\rotatebox{75}{m4\_hourly/H} & 
\rotatebox{75}{m4\_monthly/M} & 
\rotatebox{75}{m4\_quarterly/Q} & 
\rotatebox{75}{m4\_weekly/W} & 
\rotatebox{75}{m4\_yearly/A} & 
\rotatebox{75}{m\_dense/D} & 
\rotatebox{75}{m\_dense/H} & 
\rotatebox{75}{restaurant/D} & 
\rotatebox{75}{saugeen/D} & 
\rotatebox{75}{saugeen/M} & 
\rotatebox{75}{saugeen/W} & 
\rotatebox{75}{solar/10T} & 
\rotatebox{75}{solar/D} & 
\rotatebox{75}{solar/H} & 
\rotatebox{75}{solar/W} & 
\rotatebox{75}{sz\_taxi/15T} & 
\rotatebox{75}{sz\_taxi/H} & 
\rotatebox{75}{temperature\_rain/D} & 
\rotatebox{75}{us\_births/D} & 
\rotatebox{75}{us\_births/M} & 
\rotatebox{75}{us\_births/W} & 
\rotatebox{75}{bitbrains\_fast\_storage/5T} & 
\rotatebox{75}{bitbrains\_fast\_storage/H} & 
\rotatebox{75}{bitbrains\_rnd/5T} & 
\rotatebox{75}{bitbrains\_rnd/H} \\
\midrule
HART-36M          & 0.654 & 1.005 & 0.558 & 0.834 & 0.874 & 0.703 & 1.100 & 0.407 & 0.520 & 0.671 & 0.824 & 0.797 & 0.566 & 0.991 & 0.841 & 0.867 & 0.675 & 0.717 & 0.770 & 0.656 & 0.193 & 1.243 & 0.731 & 0.589 & 0.827 & 0.828 & 0.970 \\
RWKV-36M          & 0.665 & 0.996 & 0.550 & 0.824 & 0.865 & 0.698 & 1.112 & 0.412 & 0.512 & 0.672 & 0.789 & 0.778 & 0.562 & 0.797 & 0.844 & 0.878 & 0.670 & 0.717 & 0.770 & 0.710 & 0.202 & 1.541 & 0.652 & 0.591 & 0.830 & 0.829 & 0.969 \\
Transformer-35M   & 0.664 & 0.998 & 0.554 & 0.826 & 0.860 & 0.706 & 1.097 & 0.396 & 0.535 & 0.674 & 0.846 & 0.808 & 0.593 & 0.913 & 0.838 & 0.884 & 0.761 & 0.718 & 0.773 & 0.656 & 0.183 & 1.458 & 0.676 & 0.595 & 0.826 & 0.833 & 0.970 \\
Kairos-23M        & 0.667 & 1.011 & 0.882 & 0.771 & 0.743 & 0.892 & 0.752 & 0.407 & 0.571 & 0.692 & 0.895 & 0.715 & 0.623 & 1.071 & 0.859 & 0.912 & 0.715 & 0.727 & 0.773 & 0.699 & 0.234 & 1.100 & 0.794 & 0.711 & 0.854 & 0.889 & 0.980 \\
TiRex-35M         & 0.638 & 0.986 & 0.605 & 0.731 & 0.715 & 0.717 & 0.848 & 0.405 & 0.526 & 0.674 & 0.879 & 0.767 & 0.588 & 0.984 & 0.827 & 0.782 & 0.707 & 0.710 & 0.763 & 0.660 & 0.237 & 0.791 & 0.669 & 0.608 & 0.785 & 0.840 & 0.963 \\
Toto-2.0-22M      & 0.680 & 0.978 & 0.691 & 0.728 & 0.730 & 0.728 & 0.826 & 0.415 & 0.539 & 0.700 & 0.878 & 0.732 & 0.630 & 0.822 & 0.842 & 0.912 & 0.827 & 0.715 & 0.765 & 0.682 & 0.208 & 0.927 & 0.695 & 0.557 & 0.707 & 0.817 & 0.959 \\
Toto-2.0-313M     & 0.638 & 0.970 & 0.583 & 0.731 & 0.717 & 0.671 & 0.848 & 0.395 & 0.509 & 0.689 & 0.881 & 0.731 & 0.604 & 0.776 & 0.835 & 0.924 & 0.750 & 0.709 & 0.759 & 0.678 & 0.185 & 0.914 & 0.579 & 0.562 & 0.718 & 0.800 & 0.956 \\
Toto-2.0-1B       & 0.628 & 0.974 & 0.560 & 0.729 & 0.715 & 0.683 & 0.856 & 0.395 & 0.510 & 0.689 & 0.882 & 0.734 & 0.617 & 0.790 & 0.831 & 0.976 & 0.771 & 0.707 & 0.759 & 0.679 & 0.188 & 0.949 & 0.594 & 0.554 & 0.708 & 0.796 & 0.955 \\
TempoPFN-40M      & 0.730 & 1.334 & 0.678 & 0.732 & 0.723 & 0.898 & 0.906 & 0.429 & 0.611 & 0.682 & 0.926 & 0.786 & 0.629 & 0.896 & 0.844 & 0.912 & 0.905 & 0.727 & 0.776 & 0.678 & 0.207 & 0.815 & 0.583 & 0.705 & 0.907 & 0.930 & 0.996 \\
Kairos-50M        & 0.642 & 1.010 & 0.854 & 0.773 & 0.753 & 0.876 & 0.768 & 0.402 & 0.563 & 0.689 & 0.815 & 0.720 & 0.608 & 1.011 & 0.861 & 0.887 & 0.689 & 0.720 & 0.769 & 0.687 & 0.215 & 0.994 & 0.764 & 0.684 & 0.837 & 0.872 & 0.951 \\
Sundial-128M      & 0.681 & 1.133 & 0.729 & 0.873 & 0.910 & 0.863 & 1.092 & 0.408 & 0.532 & 0.700 & 0.816 & 0.771 & 0.602 & 0.757 & 0.936 & 0.827 & 0.667 & 0.724 & 0.788 & 0.711 & 0.208 & 1.523 & 0.871 & 0.652 & 0.886 & 0.870 & 0.990 \\
YingLong-110M     & 0.685 & 1.081 & 0.808 & 0.861 & 0.920 & 0.833 & 1.189 & 0.520 & 0.637 & 0.706 & 0.881 & 0.814 & 0.561 & 1.031 & 0.842 & 0.925 & 1.478 & 0.715 & 0.765 & 0.701 & 0.307 & 0.855 & 0.831 & 0.682 & 0.851 & 0.892 & 0.980 \\
Moirai2-12.8M     & 0.658 & 0.937 & 0.678 & 0.746 & 0.739 & 0.765 & 0.837 & 0.414 & 0.533 & 0.692 & 0.792 & 0.749 & 0.705 & 0.595 & 0.920 & 0.924 & 0.814 & 0.715 & 0.763 & 0.670 & 0.200 & 1.067 & 0.557 & 0.604 & 0.822 & 0.838 & 0.962 \\
Xihe-1.5B         & 0.636 & 1.001 & 0.649 & 0.738 & 0.745 & 0.774 & 0.839 & 0.428 & 0.528 & 0.676 & 0.890 & 0.774 & 0.612 & 0.911 & 0.837 & 0.878 & 0.656 & 0.717 & 0.763 & 0.682 & 0.203 & 1.125 & 0.808 & 0.635 & 0.835 & 0.855 & 0.968 \\
TimesFM-2.0-500M  & 0.643 & 0.941 & 0.499 & 0.477 & 0.602 & 0.800 & 0.640 & 0.381 & 0.570 & 0.688 & 0.978 & 0.857 & 0.982 & 1.336 & 0.840 & 1.073 & 0.872 & 0.705 & 0.758 & 0.713 & 0.198 & 0.653 & 0.705 & 0.644 & 0.843 & 0.898 & 0.992 \\
\bottomrule
\end{tabular}%
}
\end{table*}

\begin{table*}[htbp]
\centering
\caption{Per-dataset MASE on TIME-Benchmark long-term tasks (24 tasks).}
\label{tab:time-long-detail}
\tiny
\setlength{\tabcolsep}{1.5pt}
\resizebox{\textwidth}{!}{%
\begin{tabular}{@{}l*{24}{c}@{}}
\toprule
Model & 
\rotatebox{75}{Australia\_Solar/H} & 
\rotatebox{75}{CPHL/15T} & 
\rotatebox{75}{CPHL/30T} & 
\rotatebox{75}{CPHL/H} & 
\rotatebox{75}{Coastal\_T\_S/15T} & 
\rotatebox{75}{Coastal\_T\_S/20T} & 
\rotatebox{75}{Coastal\_T\_S/5T} & 
\rotatebox{75}{Coastal\_T\_S/H} & 
\rotatebox{75}{EWELD\_Load/15T} & 
\rotatebox{75}{Finland\_Traffic/15T} & 
\rotatebox{75}{MetroPT-3/5T} & 
\rotatebox{75}{NE\_China\_Wind/H} & 
\rotatebox{75}{OpenElectricity\_NEM/5T} & 
\rotatebox{75}{SG\_Carpark/15T} & 
\rotatebox{75}{SG\_PM25/H} & 
\rotatebox{75}{SG\_Weather/D} & 
\rotatebox{75}{Smart\_Manufacturing/H} & 
\rotatebox{75}{Water\_Quality\_Darwin/15T} & 
\rotatebox{75}{current\_velocity/10T} & 
\rotatebox{75}{current\_velocity/15T} & 
\rotatebox{75}{current\_velocity/20T} & 
\rotatebox{75}{current\_velocity/5T} & 
\rotatebox{75}{current\_velocity/H} & 
\rotatebox{75}{epf\_electricity\_price/H} \\
\midrule
HART-36M          & 0.834 & 0.779 & 0.806 & 0.805 & 0.860 & 0.796 & 0.813 & 0.885 & 0.664 & 0.479 & 0.748 & 0.725 & 0.690 & 0.492 & 0.696 & 0.710 & 0.666 & 0.869 & 0.667 & 0.752 & 0.790 & 0.663 & 0.671 & 0.687 \\
RWKV-36M          & 0.847 & 0.764 & 0.846 & 0.815 & 0.857 & 0.881 & 0.858 & 0.868 & 0.723 & 0.595 & 0.759 & 0.713 & 0.766 & 0.598 & 0.700 & 0.710 & 0.668 & 0.848 & 0.718 & 0.765 & 0.805 & 0.794 & 0.655 & 0.680 \\
Transformer-35M   & 0.861 & 0.738 & 0.788 & 0.798 & 0.853 & 0.810 & 0.902 & 0.839 & 0.664 & 0.491 & 0.749 & 0.744 & 0.669 & 0.542 & 0.711 & 0.719 & 0.666 & 0.832 & 0.689 & 0.759 & 0.796 & 0.787 & 0.694 & 0.694 \\
TiRex-35M         & 0.830 & 0.751 & 0.804 & 0.858 & 0.846 & 0.837 & 0.852 & 0.885 & 0.715 & 0.568 & 0.743 & 0.796 & 0.862 & 0.623 & 0.705 & 0.715 & 0.667 & 0.905 & 0.729 & 0.763 & 0.809 & 0.737 & 0.698 & 0.679 \\
Toto-2.0-22M      & 0.890 & 0.750 & 0.812 & 0.846 & 0.854 & 0.761 & 0.820 & 0.802 & 0.718 & 0.520 & 0.751 & 0.844 & 0.727 & 0.624 & 0.693 & 0.711 & 0.669 & 0.811 & 0.676 & 0.735 & 0.754 & 0.709 & 0.662 & 0.704 \\
Toto-2.0-313M     & 0.835 & 0.730 & 0.792 & 0.826 & 0.846 & 0.662 & 0.811 & 0.771 & 0.663 & 0.467 & 0.801 & 0.803 & 0.701 & 0.567 & 0.684 & 0.705 & 0.669 & 0.787 & 0.619 & 0.723 & 0.726 & 0.603 & 0.633 & 0.681 \\
Moirai2-12.8M     & 0.905 & 0.822 & 0.824 & 0.839 & 0.873 & 0.911 & 0.864 & 0.846 & 0.759 & 0.501 & 0.808 & 0.842 & 0.748 & 0.601 & 0.720 & 0.727 & 0.666 & 0.972 & 0.895 & 0.799 & 0.900 & 1.047 & 0.699 & 0.711 \\
Kairos-50M        & 0.848 & 0.782 & 0.834 & 0.835 & 0.848 & 0.992 & 0.943 & 0.875 & 0.723 & 0.504 & 0.782 & 0.826 & 0.757 & 0.599 & 0.707 & 0.725 & 0.665 & 1.016 & 0.855 & 0.908 & 0.868 & 1.016 & 0.812 & 0.753 \\
Sundial-128M      & 0.936 & 0.721 & 0.806 & 0.917 & 0.869 & 1.304 & 1.001 & 1.035 & 0.734 & 0.583 & 0.799 & 0.773 & 0.804 & 0.614 & 0.706 & 0.736 & 0.696 & 1.159 & 0.851 & 0.835 & 0.921 & 0.706 & 1.146 & 0.688 \\
TS-ICL            & 0.846 & 0.763 & 0.797 & 0.787 & 0.859 & 0.961 & 0.895 & 0.850 & 0.730 & 0.495 & 0.765 & 0.810 & 0.757 & 0.612 & 0.704 & 0.710 & 0.664 & 0.911 & 0.730 & 0.759 & 0.829 & 0.951 & 0.662 & 0.700 \\
TimesFM-2.0-500M  & 0.930 & 0.739 & 0.795 & 0.835 & 0.879 & 1.088 & 0.868 & 0.955 & 0.708 & 0.555 & 0.863 & 0.805 & 0.951 & 0.636 & 0.708 & 0.740 & 0.674 & 1.015 & 0.735 & 0.813 & 0.810 & 0.702 & 0.814 & 0.723 \\
\bottomrule
\end{tabular}%
}
\end{table*}

\begin{table*}[htbp]
\centering
\caption{Per-dataset MASE on TIME-Benchmark medium-term tasks (24 tasks).}
\label{tab:time-medium-detail}
\tiny
\setlength{\tabcolsep}{1.5pt}
\resizebox{\textwidth}{!}{%
\begin{tabular}{@{}l*{24}{c}@{}}
\toprule
Model & 
\rotatebox{75}{Australia\_Solar/H} & 
\rotatebox{75}{CPHL/15T} & 
\rotatebox{75}{CPHL/30T} & 
\rotatebox{75}{CPHL/H} & 
\rotatebox{75}{Coastal\_T\_S/15T} & 
\rotatebox{75}{Coastal\_T\_S/20T} & 
\rotatebox{75}{Coastal\_T\_S/5T} & 
\rotatebox{75}{Coastal\_T\_S/H} & 
\rotatebox{75}{EWELD\_Load/15T} & 
\rotatebox{75}{Finland\_Traffic/15T} & 
\rotatebox{75}{MetroPT-3/5T} & 
\rotatebox{75}{NE\_China\_Wind/H} & 
\rotatebox{75}{OpenElectricity\_NEM/5T} & 
\rotatebox{75}{SG\_Carpark/15T} & 
\rotatebox{75}{SG\_PM25/H} & 
\rotatebox{75}{SG\_Weather/D} & 
\rotatebox{75}{Smart\_Manufacturing/H} & 
\rotatebox{75}{Water\_Quality\_Darwin/15T} & 
\rotatebox{75}{current\_velocity/10T} & 
\rotatebox{75}{current\_velocity/15T} & 
\rotatebox{75}{current\_velocity/20T} & 
\rotatebox{75}{current\_velocity/5T} & 
\rotatebox{75}{current\_velocity/H} & 
\rotatebox{75}{epf\_electricity\_price/H} \\
\midrule
HART-36M          & 0.773 & 0.758 & 0.813 & 0.808 & 0.843 & 0.780 & 0.774 & 0.799 & 0.666 & 0.682 & 0.755 & 0.750 & 0.532 & 0.534 & 0.758 & 0.729 & 0.683 & 0.834 & 0.667 & 0.725 & 0.740 & 0.646 & 0.624 & 0.642 \\
RWKV-36M          & 0.787 & 0.765 & 0.824 & 0.800 & 0.847 & 0.780 & 0.776 & 0.825 & 0.685 & 0.706 & 0.760 & 0.725 & 0.610 & 0.588 & 0.762 & 0.731 & 0.686 & 0.804 & 0.698 & 0.740 & 0.739 & 0.646 & 0.607 & 0.639 \\
Transformer-35M   & 0.787 & 0.733 & 0.820 & 0.792 & 0.841 & 0.739 & 0.799 & 0.814 & 0.674 & 0.707 & 0.754 & 0.768 & 0.530 & 0.568 & 0.770 & 0.735 & 0.684 & 0.824 & 0.680 & 0.731 & 0.744 & 0.702 & 0.641 & 0.662 \\
TiRex-35M         & 0.805 & 0.758 & 0.834 & 0.826 & 0.851 & 0.837 & 0.805 & 0.857 & 0.694 & 0.811 & 0.743 & 0.789 & 0.689 & 0.703 & 0.766 & 0.735 & 0.685 & 0.876 & 0.725 & 0.736 & 0.748 & 0.750 & 0.636 & 0.629 \\
Toto-2.0-22M      & 0.826 & 0.744 & 0.804 & 0.847 & 0.848 & 0.694 & 0.784 & 0.743 & 0.681 & 0.762 & 0.749 & 0.811 & 0.583 & 0.718 & 0.756 & 0.731 & 0.687 & 0.762 & 0.681 & 0.724 & 0.715 & 0.663 & 0.616 & 0.655 \\
Toto-2.0-313M     & 0.771 & 0.751 & 0.802 & 0.832 & 0.844 & 0.659 & 0.772 & 0.719 & 0.651 & 0.646 & 0.778 & 0.794 & 0.523 & 0.564 & 0.742 & 0.726 & 0.684 & 0.734 & 0.645 & 0.728 & 0.695 & 0.639 & 0.593 & 0.632 \\
Moirai2-12.8M     & 0.848 & 0.756 & 0.867 & 0.819 & 0.855 & 0.914 & 0.838 & 0.781 & 0.707 & 0.687 & 0.814 & 0.818 & 0.599 & 0.593 & 0.782 & 0.746 & 0.684 & 0.922 & 0.847 & 0.767 & 0.821 & 0.833 & 0.657 & 0.655 \\
Kairos-50M        & 0.819 & 0.793 & 0.846 & 0.817 & 0.849 & 1.061 & 0.865 & 0.839 & 0.729 & 0.721 & 0.802 & 0.793 & 0.603 & 0.674 & 0.767 & 0.745 & 0.682 & 1.084 & 0.919 & 0.851 & 0.853 & 0.841 & 0.832 & 0.687 \\
Sundial-128M      & 0.852 & 0.722 & 0.836 & 0.907 & 0.851 & 1.206 & 0.889 & 0.956 & 0.711 & 0.725 & 0.807 & 0.840 & 0.629 & 0.607 & 0.768 & 0.751 & 0.699 & 1.144 & 0.852 & 0.802 & 0.866 & 0.726 & 0.994 & 0.660 \\
TS-ICL            & 0.797 & 0.789 & 0.827 & 0.798 & 0.850 & 0.968 & 0.805 & 0.772 & 0.687 & 0.657 & 0.767 & 0.794 & 0.613 & 0.641 & 0.765 & 0.733 & 0.682 & 0.905 & 0.733 & 0.754 & 0.789 & 0.702 & 0.629 & 0.641 \\
TimesFM-2.0-500M  & 0.871 & 0.741 & 0.850 & 0.833 & 0.861 & 1.021 & 0.825 & 0.930 & 0.712 & 0.783 & 0.820 & 0.806 & 0.750 & 0.664 & 0.772 & 0.758 & 0.683 & 1.061 & 0.742 & 0.768 & 0.776 & 0.691 & 0.728 & 0.667 \\
\bottomrule
\end{tabular}%
}
\end{table*}

\begin{table*}[htbp]
\centering
\caption{Per-dataset MASE on TIME-Benchmark short-term tasks (Part 1: A to M, 25 tasks).}
\label{tab:time-short-detail-1}
\tiny
\setlength{\tabcolsep}{1.5pt}
\resizebox{\textwidth}{!}{%
\begin{tabular}{@{}l*{25}{c}@{}}
\toprule
Model & 
\rotatebox{75}{Australia\_Solar/H} & 
\rotatebox{75}{Auto\_Production\_SF/M} & 
\rotatebox{75}{CPHL/15T} & 
\rotatebox{75}{CPHL/30T} & 
\rotatebox{75}{CPHL/H} & 
\rotatebox{75}{Coastal\_T\_S/15T} & 
\rotatebox{75}{Coastal\_T\_S/20T} & 
\rotatebox{75}{Coastal\_T\_S/5T} & 
\rotatebox{75}{Coastal\_T\_S/H} & 
\rotatebox{75}{Commodity\_Import/M} & 
\rotatebox{75}{Commodity\_Production/M} & 
\rotatebox{75}{Crypto/D} & 
\rotatebox{75}{ECDC\_COVID/D} & 
\rotatebox{75}{ECDC\_COVID/W} & 
\rotatebox{75}{EWELD\_Load/15T} & 
\rotatebox{75}{Finland\_Traffic/15T} & 
\rotatebox{75}{Global\_Influenza/W} & 
\rotatebox{75}{Global\_Price/Q} & 
\rotatebox{75}{Housing\_Inventory/M} & 
\rotatebox{75}{JOLTS/M} & 
\rotatebox{75}{Job\_Claims/W} & 
\rotatebox{75}{MetroPT-3/5T} & 
\rotatebox{75}{NE\_China\_Wind/H} & 
\rotatebox{75}{Oil\_Price/B} & 
\rotatebox{75}{Online\_Retail\_2\_UCI/D} \\
\midrule
HART-36M          & 0.811 & 0.797 & 0.457 & 0.710 & 0.796 & 0.758 & 0.332 & 0.619 & 0.722 & 0.831 & 0.787 & 1.016 & 0.975 & 0.882 & 0.548 & 0.497 & 0.601 & 0.804 & 0.934 & 0.653 & 1.070 & 0.708 & 0.559 & 0.887 & 0.271 \\
RWKV-36M          & 0.815 & 0.719 & 0.466 & 0.717 & 0.798 & 0.760 & 0.328 & 0.617 & 0.724 & 0.818 & 0.768 & 1.003 & 0.983 & 0.904 & 0.554 & 0.475 & 0.598 & 0.806 & 0.957 & 0.657 & 1.044 & 0.713 & 0.572 & 0.902 & 0.291 \\
Transformer-35M   & 0.797 & 0.841 & 0.464 & 0.712 & 0.805 & 0.760 & 0.324 & 0.621 & 0.726 & 0.802 & 0.781 & 1.019 & 0.893 & 0.955 & 0.553 & 0.469 & 0.615 & 0.801 & 0.909 & 0.627 & 1.015 & 0.701 & 0.574 & 0.870 & 0.271 \\
TiRex-35M         & 0.839 & 0.763 & 0.470 & 0.727 & 0.804 & 0.762 & 0.342 & 0.626 & 0.729 & 0.802 & 0.733 & 0.985 & 0.815 & 0.808 & 0.578 & 0.586 & 0.649 & 0.820 & 0.672 & 0.609 & 0.854 & 0.702 & 0.576 & 0.889 & 0.284 \\
Toto-2.0-22M      & 0.892 & 0.698 & 0.468 & 0.704 & 0.790 & 0.760 & 0.319 & 0.624 & 0.719 & 0.802 & 0.742 & 1.052 & 0.847 & 0.996 & 0.566 & 0.532 & 0.679 & 0.771 & 0.707 & 0.571 & 0.873 & 0.735 & 0.600 & 0.877 & 0.311 \\
Toto-2.0-313M     & 0.781 & 0.600 & 0.463 & 0.700 & 0.792 & 0.757 & 0.292 & 0.616 & 0.704 & 0.783 & 0.706 & 1.143 & 0.807 & 0.985 & 0.543 & 0.440 & 0.650 & 0.780 & 0.591 & 0.517 & 0.862 & 0.723 & 0.583 & 0.865 & 0.292 \\
Moirai2-12.8M     & 0.868 & 0.926 & 0.473 & 0.762 & 0.815 & 0.764 & 0.438 & 0.641 & 0.748 & 0.779 & 0.763 & 1.030 & 0.948 & 0.905 & 0.592 & 0.505 & 0.604 & 0.810 & 0.732 & 0.637 & 0.984 & 0.741 & 0.593 & 0.921 & 0.294 \\
Kairos-50M        & 0.843 & 1.257 & 0.477 & 0.739 & 0.825 & 0.769 & 0.504 & 0.652 & 0.815 & 0.833 & 0.744 & 1.235 & 1.102 & 1.439 & 0.636 & 0.525 & 0.585 & 0.829 & 0.712 & 0.761 & 1.188 & 0.782 & 0.596 & 0.895 & 0.305 \\
Sundial-128M      & 0.858 & 1.435 & 0.465 & 0.731 & 0.842 & 0.763 & 0.422 & 0.643 & 0.801 & 0.971 & 0.967 & 1.022 & 0.888 & 0.801 & 0.585 & 0.485 & 0.739 & 0.958 & 1.440 & 0.946 & 1.959 & 0.771 & 0.607 & 0.862 & 0.357 \\
TS-ICL            & 0.823 & 0.904 & 0.474 & 0.717 & 0.787 & 0.769 & 0.444 & 0.636 & 0.758 & 0.764 & 0.756 & 1.016 & 0.909 & 1.139 & 0.577 & 0.525 & 0.693 & 0.846 & 0.723 & 0.652 & 0.992 & 0.726 & 0.591 & 0.866 & 0.301 \\
TimesFM-2.0-500M  & 0.950 & 1.015 & 0.473 & 0.730 & 0.813 & 0.765 & 0.398 & 0.634 & 0.790 & 0.899 & 0.637 & 1.084 & 0.957 & 1.160 & 0.617 & 0.592 & 0.579 & 0.821 & 0.787 & 0.679 & 1.019 & 0.770 & 0.584 & 0.896 & 0.300 \\
\bottomrule
\end{tabular}%
}
\end{table*}

\begin{table*}[htbp]
\centering
\caption{Per-dataset MASE on TIME-Benchmark short-term tasks (Part 2: O--Z, 25 tasks).}
\label{tab:time-short-detail-2}
\tiny
\setlength{\tabcolsep}{1.5pt}
\resizebox{\textwidth}{!}{%
\begin{tabular}{@{}l*{25}{c}@{}}
\toprule
Model & 
\rotatebox{75}{OpenElectricity\_NEM/5T} & 
\rotatebox{75}{Port\_Activity/D} & 
\rotatebox{75}{Port\_Activity/W} & 
\rotatebox{75}{SG\_Carpark/15T} & 
\rotatebox{75}{SG\_PM25/H} & 
\rotatebox{75}{SG\_Weather/D} & 
\rotatebox{75}{Smart\_Manufacturing/H} & 
\rotatebox{75}{Supply\_Chain\_Customer/D} & 
\rotatebox{75}{Supply\_Chain\_Location/D} & 
\rotatebox{75}{US\_Labor/M} & 
\rotatebox{75}{US\_Term\_Structure/B} & 
\rotatebox{75}{Uncertainty\_1M/M} & 
\rotatebox{75}{Vehicle\_Sales/M} & 
\rotatebox{75}{Vehicle\_Supply/M} & 
\rotatebox{75}{WUI\_Global/Q} & 
\rotatebox{75}{Water\_Quality\_Darwin/15T} & 
\rotatebox{75}{azure2019\_D/5T} & 
\rotatebox{75}{azure2019\_I/5T} & 
\rotatebox{75}{azure2019\_U/5T} & 
\rotatebox{75}{current\_velocity/10T} & 
\rotatebox{75}{current\_velocity/15T} & 
\rotatebox{75}{current\_velocity/20T} & 
\rotatebox{75}{current\_velocity/5T} & 
\rotatebox{75}{current\_velocity/H} & 
\rotatebox{75}{epf\_electricity\_price/H} \\
\midrule
HART-36M          & 0.364 & 0.578 & 0.766 & 0.372 & 0.736 & 0.779 & 0.710 & 0.340 & 0.393 & 0.414 & 0.872 & 0.318 & 0.581 & 0.794 & 0.720 & 0.559 & 0.686 & 0.688 & 0.601 & 0.494 & 0.499 & 0.464 & 0.396 & 0.663 & 0.669 \\
RWKV-36M          & 0.385 & 0.577 & 0.769 & 0.389 & 0.739 & 0.779 & 0.714 & 0.343 & 0.397 & 0.438 & 0.846 & 0.338 & 0.588 & 0.798 & 0.716 & 0.545 & 0.669 & 0.680 & 0.595 & 0.513 & 0.501 & 0.463 & 0.391 & 0.651 & 0.665 \\
Transformer-35M   & 0.361 & 0.576 & 0.769 & 0.380 & 0.742 & 0.786 & 0.712 & 0.338 & 0.389 & 0.416 & 0.863 & 0.319 & 0.576 & 0.797 & 0.721 & 0.556 & 0.689 & 0.696 & 0.604 & 0.500 & 0.501 & 0.464 & 0.423 & 0.665 & 0.687 \\
TiRex-35M         & 0.414 & 0.576 & 0.753 & 0.434 & 0.743 & 0.780 & 0.713 & 0.356 & 0.406 & 0.443 & 0.849 & 0.351 & 0.570 & 0.714 & 0.721 & 0.567 & 0.697 & 0.719 & 0.581 & 0.512 & 0.500 & 0.467 & 0.408 & 0.654 & 0.658 \\
Toto-2.0-22M      & 0.374 & 0.574 & 0.756 & 0.432 & 0.732 & 0.776 & 0.728 & 0.337 & 0.389 & 0.441 & 0.935 & 0.358 & 0.536 & 0.704 & 0.715 & 0.517 & 0.658 & 0.670 & 0.571 & 0.515 & 0.501 & 0.458 & 0.406 & 0.665 & 0.676 \\
Toto-2.0-313M     & 0.354 & 0.573 & 0.754 & 0.379 & 0.725 & 0.775 & 0.722 & 0.324 & 0.379 & 0.451 & 0.869 & 0.370 & 0.516 & 0.690 & 0.713 & 0.493 & 0.619 & 0.623 & 0.549 & 0.505 & 0.495 & 0.446 & 0.423 & 0.642 & 0.659 \\
Moirai2-12.8M     & 0.402 & 0.575 & 0.749 & 0.411 & 0.756 & 0.790 & 0.710 & 0.353 & 0.406 & 0.454 & 0.886 & 0.354 & 0.538 & 0.682 & 0.725 & 0.606 & 0.693 & 0.718 & 0.588 & 0.561 & 0.518 & 0.512 & 0.420 & 0.698 & 0.671 \\
Kairos-50M        & 0.415 & 0.578 & 0.763 & 0.449 & 0.742 & 0.795 & 0.710 & 0.363 & 0.407 & 0.592 & 0.920 & 0.321 & 0.665 & 0.796 & 0.729 & 0.725 & 0.762 & 0.780 & 0.662 & 0.619 & 0.554 & 0.539 & 0.460 & 0.892 & 0.706 \\
Sundial-128M      & 0.399 & 0.589 & 0.777 & 0.405 & 0.743 & 0.794 & 0.723 & 0.393 & 0.441 & 0.813 & 0.867 & 0.399 & 0.641 & 0.646 & 0.763 & 0.652 & 0.709 & 0.753 & 0.680 & 0.554 & 0.544 & 0.546 & 0.413 & 1.112 & 0.693 \\
TS-ICL            & 0.410 & 0.579 & 0.765 & 0.416 & 0.741 & 0.777 & 0.709 & 0.360 & 0.413 & 0.498 & 0.858 & 0.328 & 0.564 & 0.714 & 0.720 & 0.609 & 0.686 & 0.713 & 0.604 & 0.555 & 0.517 & 0.508 & 0.436 & 0.703 & 0.664 \\
TimesFM-2.0-500M  & 0.432 & 0.577 & 0.742 & 0.436 & 0.750 & 0.799 & 0.712 & 0.362 & 0.406 & 0.467 & 0.865 & 0.424 & 0.626 & 0.639 & 0.725 & 0.650 & 0.701 & 0.773 & 0.626 & 0.531 & 0.515 & 0.486 & 0.409 & 0.740 & 0.690 \\
\bottomrule
\end{tabular}%
}
\end{table*}

\subsection{GIFT-Eval}

Table~\ref{tab:long-detail} reports MASE on all 21 long-term datasets.
Long-term tasks use prediction lengths between 720 and 900 time steps.

Table~\ref{tab:medium-detail} reports MASE on all 21 medium-term datasets.
Medium-term tasks use prediction lengths of approximately 500 time steps.

Table~\ref{tab:short-detail-1} and Table~\ref{tab:short-detail-2} report MASE on all 55 short-term datasets.
Short-term tasks use prediction lengths of approximately 30 time steps.

\subsection{TIME-Benchmark}

Table~\ref{tab:time-long-detail} reports MASE on all 24 long-term tasks.
Table~\ref{tab:time-medium-detail} reports MASE on all 24 medium-term tasks.
Table~\ref{tab:time-short-detail-1} and Table~\ref{tab:time-short-detail-2} report MASE on all 50 short-term tasks.

\section{Mechanistic Evidence of PID}\label{sec:pid-mechanism}

This section provides the full experimental setup and temporal evolution evidence for the norm analysis in Section~\ref{sec:mechanistic-evidence}. The table in the main text reports region-averaged ratios between the historical and prediction regions; here we present two figures showing the temporal evolution of the corresponding norms near the history--prediction boundary.

\subsection{Experimental Setup}\label{sec:pid-mech-setup}

The synthetic signal consists of three consecutive sinusoidal segments designed to mimic structural changes commonly found in real-world time series. The three segments have periods of 100, 250, and 80 time steps, with amplitudes sampled from $\mathcal{U}(0.8,1.2)$, $\mathcal{U}(1.0,1.5)$, and $\mathcal{U}(0.6,1.0)$, respectively, and phases sampled from $\mathcal{U}(0,2\pi)$. Gaussian noise with a standard deviation of 0.1 is added to the entire signal. Each sequence has a total length of 17,408 time steps. With patch size $P=16$, the historical region contains 1,024 patches and the prediction region contains 64 patches, yielding 1,088 patches in total.

Inference follows the NAF paradigm: the input is the concatenation of historical patches and prediction placeholder patches, and all prediction steps are produced in a single forward pass. We generate 4,096 independent sequences, each with randomly sampled phases and amplitudes. During layer-by-layer forward propagation, we record $\|K_t\|_F$ and $\|M_t\|_F$ at every patch position, average them across the 8 heads, and then average across all 4,096 sequences, yielding a norm matrix of dimension 12 layers $\times$ 1,088 patches. The visualization focuses on the boundary region: we display the last 64 historical patches and all 64 prediction patches, with a red dashed line marking the boundary.

\subsection{Temporal Evolution of \texorpdfstring{$K_t$}{Kt} and \texorpdfstring{$M_t$}{Mt}}

\begin{figure*}[t]
    \centering
    \includegraphics[width=\textwidth]{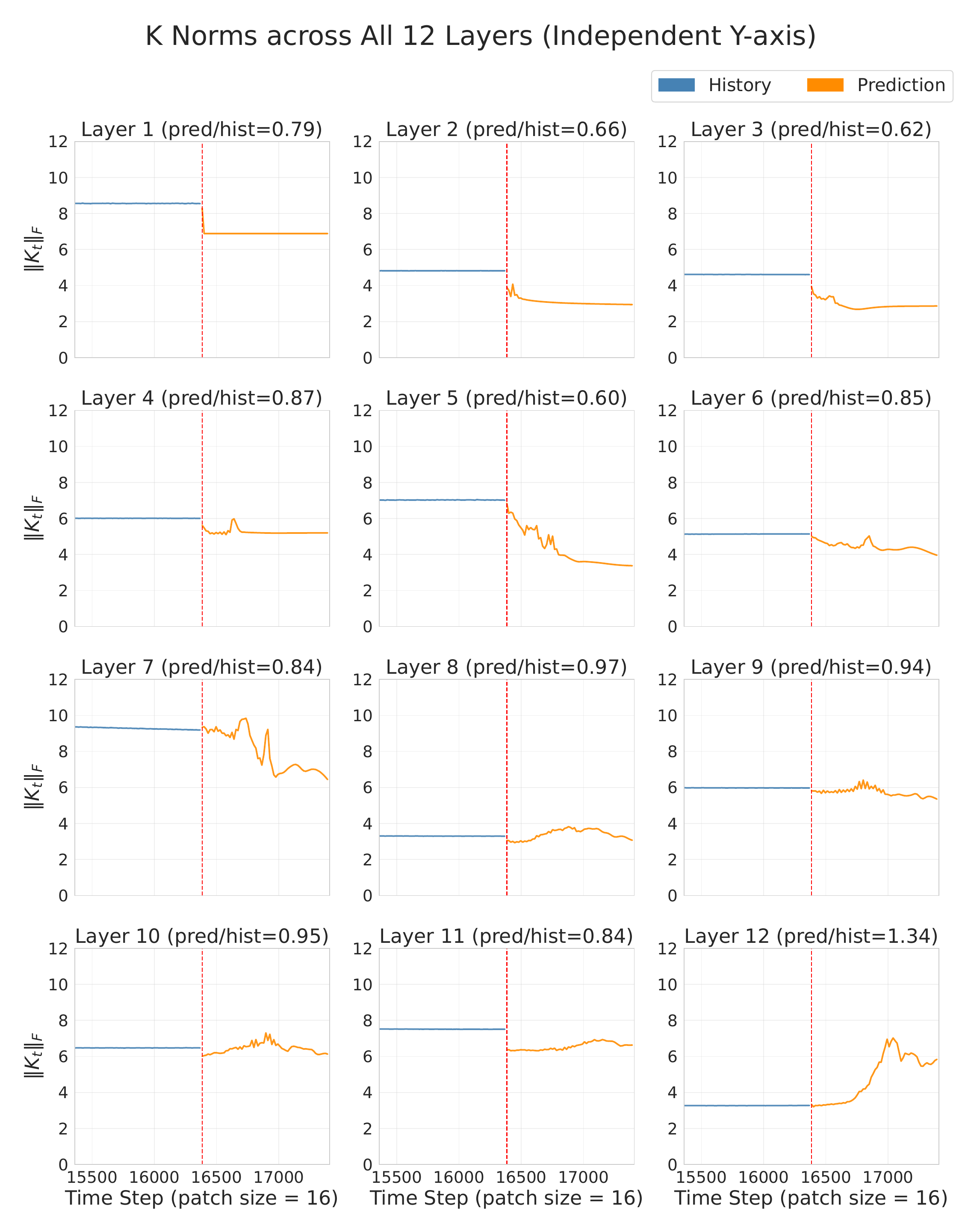}
    \caption{Temporal evolution of $\|K_t\|_F$ across all 12 layers. Blue: historical region; orange: prediction region; red dashed line marks the boundary. Prediction-region averages are lower in Layers 1 to 11, with transient increases in some layers; Layer 12 has a higher prediction-region average.}
    \label{fig:pid-k-evolution}
\end{figure*}

\begin{figure*}[t]
    \centering
    \includegraphics[width=\textwidth]{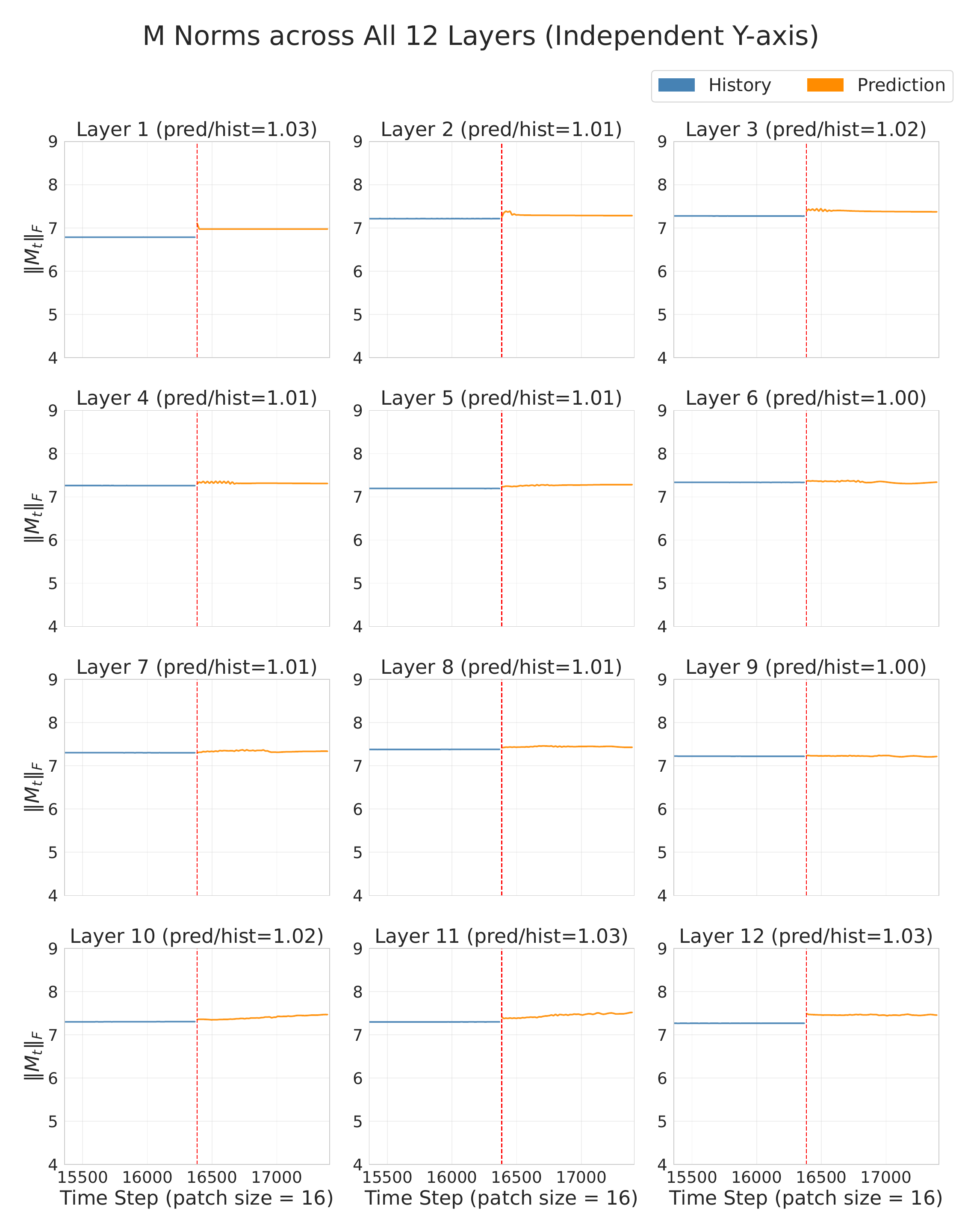}
    \caption{Temporal evolution of $\|M_t\|_F$ across all 12 layers. Blue: historical region; orange: prediction region; red dashed line marks the boundary. The $M_t$ norms remain similar across the boundary and throughout the prediction region, with small increases in some layers.}
    \label{fig:pid-m-evolution}
\end{figure*}

Figures~\ref{fig:pid-k-evolution} and~\ref{fig:pid-m-evolution} show the temporal evolution of the $K_t$ and $M_t$ norms near the boundary, respectively.

\paragraph{$K_t$: lower prediction-region averages with layer-dependent dynamics.}

The prediction-to-history mean ratios are below one in Layers 1 to 11. The largest relative reductions occur in Layers 2, 3, and 5, with ratios from 0.60 to 0.66; Layers 8 to 11 have ratios from 0.84 to 0.97.
The temporal traces show that this regional pattern is not a uniform pointwise decrease: the earlier layers generally maintain lower norms, while several later layers show transient recoveries toward or above the historical level.
Thus, the region-averaged reduction should be distinguished from the within-region dynamics.

Layer 12 has a higher prediction-region mean norm, rising from approximately 3.2 to 7.2 before settling around 5.8.
This reversal highlights variation across depth. The norms alone do not establish whether this layer restores information affected by preceding updates.

\paragraph{$M_t$: similar transition-matrix norms.}

The $M_t$ norm is nearly constant within the displayed historical region across all 12 layers. After the boundary, it remains similar or increases slightly; Layers 10 to 12 rise gradually from about 7.3 to 7.5.
Figure~\ref{fig:pid-m-evolution} shows that these small norm differences persist through the displayed prediction interval.
They do not, by themselves, determine the contraction rate of successive transitions or the amount of historical information retained.

Together, the two figures extend the region-averaged observations in Section~\ref{sec:mechanistic-evidence} by showing their temporal structure.
Lower write norms in most layers coexist with similar transition norms, while transient recoveries and the Layer 12 reversal qualify the pattern.
These observations are consistent with PID at the level of update magnitudes, but do not directly measure net information loss.

\section{Synthetic Evidence and Teacher-Forced Comparisons}\label{sec:pid-counterfactual}

We use synthetic signals to examine how architecture rankings and sensitivity to missing future observations vary with signal structure.
The comparison pairs NAF with a teacher-forced condition that receives ground-truth future inputs, denoted AR in this appendix.
The NAF--AR gap is a diagnostic of sensitivity to the available observations, rather than a comparison with free-running autoregressive forecasting.
Full results are reported in Table~\ref{tab:pid-counterfactual}.

\subsection{Experimental Setup}\label{sec:pid-counterfactual-setup}

We design six synthetic experiments (SE1 to SE6), each corresponding to a specific signal structure. All experiments are conducted with a 16k context length: the historical length is 16,384 time steps, and the prediction length has three levels: Short (32 steps), Medium (480 steps), and Long (720 steps). Each experiment generates 256 independent sequences, with MASE as the evaluation metric. The AR control uses real future values as input in the prediction region under teacher forcing, whereas the NAF condition uses placeholders. Only the NAF condition matches the forecasting setup used for the main benchmark comparisons. All experiments reuse the pre-trained checkpoints from Section~\ref{sec:experiments}, without additional fine-tuning on the synthetic signals. The signal designs are described below.

\paragraph{SE1 (Regime Switch).}
The signal consists of three consecutive sinusoidal segments designed to mimic structural changes in real time series. The three segments have periods of 100, 250, and 80 time steps, with amplitudes sampled from $\mathcal{U}(0.8,1.2)$, $\mathcal{U}(1.0,1.5)$, and $\mathcal{U}(0.6,1.0)$, respectively, and phases sampled from $\mathcal{U}(0,2\pi)$. Gaussian noise with a standard deviation of 0.1 is added to the entire signal. Each segment is locally stationary, but the periodic patterns change across segments, and the history contains clearly learnable local structures.

\paragraph{SE2 (Ultra-Long Cycle).}
The signal is composed of a sinusoid with a dominant period of 50,000 time steps superimposed with a local high-frequency component with a period of 200 time steps, plus Gaussian noise with a standard deviation of 0.12. The dominant period far exceeds the historical length, so the model can only observe an approximately linear drift of the slow baseline; accurate prediction requires the model to maintain an estimate of this drift over a long time span and combine it with the local high-frequency component.

\paragraph{SE3 (AM + Phase Jumps).}
The signal is a fast sinusoid with a period of 50 time steps, modulated by an amplitude envelope with a period of 800 time steps, with phase jumps of magnitude $\pi/3$ introduced at fixed positions every 800 steps. The amplitude is sampled from $\mathcal{U}(0.8,1.2)$, and the noise standard deviation is 0.15. This signal requires the model to simultaneously parse fast local oscillations, slow global envelopes, and discrete phase jumps.

\paragraph{SE4 (Ultimate Mix).}
The signal is composed of multiple components: eight frequency components (periods 30, 80, 150, 300, 600, 1200, 2500, and 5000), a chirp component (period 100, drift rate 0.02), an amplitude-modulated component (period 400, modulation period 600), two structural breaks (periods 200 and 500), and a linear trend (slope sampled from $\mathcal{U}(0.0005,0.005)$). The noise standard deviation is 0.25. This is the most complex signal among the six, containing both predictable deterministic structures and non-stationary components that require online denoising.

\paragraph{SE5 (Noise Floor).}
The signal is a weak sinusoid with a period of 200 and amplitude sampled from $\mathcal{U}(0.2,0.4)$, superimposed with Gaussian noise of standard deviation 0.8. The weak periodic component is obscured by strong noise, providing a low-signal-to-noise control.

\paragraph{SE6 (Constant).}
The signal is a constant value of 100.0 plus Gaussian noise with a standard deviation of 1.0. The underlying signal level is constant, with observed fluctuations introduced by Gaussian noise.

\subsection{Signal-Dependent Architecture Performance}

The synthetic designs vary signal structure, noise, and non-stationarity, allowing us to compare the three trained architectures across these regimes.

\paragraph{RWKV on the noisy multi-component signal.}

SE4 is the most complex signal among the six. It superimposes multiple dynamically changing components: multi-frequency oscillations, chirp, amplitude modulation, structural breaks, and a linear trend, under strong noise ($\sigma=0.25$). Online denoising and non-stationary modeling are both required for this type of signal: the model needs to continuously track multiple simultaneously varying dynamic components in noise.

Results are reported in Table~\ref{tab:pid-counterfactual}. Under NAF, RWKV achieves the lowest Long MASE (1.058), compared with HART (1.148) and Transformer (1.151).
RWKV also outperforms Transformer under the teacher-forced AR condition (0.631 vs.\ 0.641).
This ranking suggests that recurrent modeling can be useful on this noisy multi-component signal, but the comparison does not isolate a particular filtering mechanism.
HART slightly improves on Transformer under NAF while remaining worse than pure RWKV on SE4.

\paragraph{Attention-based models on amplitude modulation and phase jumps.}

The SE3 signal requires structural parsing at three levels simultaneously: fast local oscillations (period 50), slow global envelope (period 800), and discrete phase jumps. The coupling of fast oscillation and slow envelope requires the model to see local details and global shape simultaneously.

Under the teacher-forced AR condition, RWKV has higher Long MASE than HART and Transformer (0.265 vs.\ 0.169 and 0.172).
Thus, the architecture ranking is already present when real future inputs are supplied.
Under NAF, HART achieves the lowest Long MASE (0.265), followed by Transformer (0.306) and RWKV (0.581).
These results favor the attention-based architectures on SE3, without identifying a unique cause for RWKV's gap.

RWKV's Long NAF--AR gap on SE3 is 0.316.
The SE3 comparison is compatible with both task-dependent architectural differences and sensitivity to placeholder inputs.

\paragraph{Complementarity and balance.}

SE4 and SE3 yield different architecture rankings: RWKV performs best under NAF on SE4, whereas HART performs best on SE3.
This variation supports examining hybrid architectures across diverse signal regimes rather than expecting universal dominance.
On the real benchmarks in Table~\ref{tab:main-result}, HART improves over the pure Transformer on aggregate MASE and over pure RWKV on Medium and Long horizons.

\subsection{Interpreting Sensitivity to Missing Future Observations}

We next examine how the NAF--AR gap changes across signal regimes and what this comparison can establish about PID.

The teacher-forced control supplies real future inputs instead of recursively generated predictions. Switching to NAF therefore changes the information available to the model as well as its input pattern. The resulting gap measures sensitivity to this change; it does not isolate PID from every other source of forecasting error.

\paragraph{Structured-signal comparisons.}

SE1 to SE3 provide three different structured signals on which to compare the effect of replacing real future inputs with placeholders.

On Long, RWKV has a larger NAF--AR gap than both other architectures on each signal: SE1, 0.060 (HART 0.001, Transformer 0.002); SE2, 0.095 (HART 0.030, Transformer 0.016); and SE3, 0.316 (HART 0.096, Transformer 0.134).
The consistent ranking indicates greater sensitivity of the evaluated RWKV model to the removal of future observations on these signals.
It is consistent with the motivation for PID analysis, but does not rule out architecture-specific sensitivity to task difficulty or changes in the input distribution.

\paragraph{Low-signal controls.}

SE5 and SE6 examine whether replacing future inputs with placeholders causes a similar loss on signals dominated by noise or a constant level.

SE5 contains a weak sinusoid under strong noise, whereas SE6 has a constant underlying level plus noise. Their limited temporal variation contrasts with the richer structures in SE1 to SE4.

The two conditions yield nearly identical scores on these controls. At the reported precision, the absolute NAF--AR gaps on SE5 are at most 0.001, while those on SE6 are zero. This shows that the loss observed on structured signals is not uniform across all tested regimes. These controls do not, by themselves, exclude overfitting, distribution shift, or other mechanisms contributing to the structured-signal gaps.

\begin{table*}[t]
\centering
\caption{NAF versus teacher-forced AR at 16k context length (MASE). AR uses real future inputs as a diagnostic control. Lower is better.}
\label{tab:pid-counterfactual}
\small
\setlength{\tabcolsep}{5pt}
\resizebox{\textwidth}{!}{%
\begin{tabular}{@{}llcccccc@{}}
\toprule
Experiment & $H$ & RWKV-AR & RWKV-NAF & HART-AR & HART-NAF & Transformer-AR & Transformer-NAF \\
\midrule
\multirow{3}{*}{SE1 Regime Switch}
 & S & 0.162 & 0.163 & 0.161 & 0.161 & 0.162 & 0.162 \\
 & M & 0.126 & 0.140 & 0.125 & 0.126 & 0.126 & 0.127 \\
 & L & 0.131 & 0.191 & 0.130 & 0.131 & 0.131 & 0.133 \\
\midrule
\multirow{3}{*}{SE2 Ultra-Long Cycle}
 & S & 0.488 & 0.490 & 0.481 & 0.483 & 0.482 & 0.483 \\
 & M & 0.205 & 0.246 & 0.200 & 0.226 & 0.199 & 0.215 \\
 & L & 0.201 & 0.296 & 0.198 & 0.228 & 0.196 & 0.212 \\
\midrule
\multirow{3}{*}{SE3 AM + Phase Jumps}
 & S & 0.267 & 0.296 & 0.164 & 0.167 & 0.169 & 0.179 \\
 & M & 0.266 & 0.480 & 0.162 & 0.255 & 0.167 & 0.313 \\
 & L & 0.265 & 0.581 & 0.169 & 0.265 & 0.172 & 0.306 \\
\midrule
\multirow{3}{*}{SE4 Ultimate Mix}
 & S & 0.695 & 0.807 & 0.766 & 0.981 & 0.831 & 1.128 \\
 & M & 0.526 & 0.914 & 0.530 & 1.029 & 0.545 & 1.044 \\
 & L & 0.631 & 1.058 & 0.635 & 1.148 & 0.641 & 1.151 \\
\midrule
\multirow{3}{*}{SE5 Noise Floor}
 & S & 0.769 & 0.769 & 0.769 & 0.769 & 0.771 & 0.771 \\
 & M & 0.751 & 0.750 & 0.751 & 0.750 & 0.751 & 0.751 \\
 & L & 0.765 & 0.764 & 0.765 & 0.764 & 0.765 & 0.765 \\
\midrule
\multirow{3}{*}{SE6 Constant}
 & S & 0.804 & 0.804 & 0.804 & 0.804 & 0.804 & 0.804 \\
 & M & 0.768 & 0.768 & 0.768 & 0.768 & 0.768 & 0.768 \\
 & L & 0.786 & 0.786 & 0.786 & 0.786 & 0.786 & 0.786 \\
\bottomrule
\end{tabular}%
}
\end{table*}

\section{Pre-training Setup}\label{sec:exp-setup}

This section details the composition and processing of pre-training data and the training hyperparameters.
All experiments are conducted under identical pre-training conditions: the three architectures, HART, pure RWKV, and pure Transformer, are trained with exactly the same data, optimization strategy, and random seed.

\subsection{Pre-training Data}

The pre-training corpus consists of real-world and synthetic time series, totaling approximately 86.7M sequences.
Roughly half are real (43.7M) and half synthetic (43M), so that the model sees both natural distributions and controlled generative signals.

Real data are drawn from three public sources:
\begin{itemize}
    \item \textbf{GIFT-Eval pre-training split}~\cite{aksu2024gift}: the same collection from which the evaluation benchmark is constructed, covering energy, transportation, meteorology, finance, and other domains. For multivariate datasets, each variate is extracted as an independent univariate series.
    \item \textbf{Chronos corpus}~\cite{ansari2025chronos}: the large-scale time series pre-training collection released by the Chronos project. Following the data-cleaning protocol of TiRex~\cite{auer2026tirex}, we remove sequences that overlap temporally with the GIFT-Eval test sets to prevent evaluation leakage.
    \item \textbf{Code15}~\cite{ribeiro2021code}: a public healthcare dataset used as an additional real-world source.
\end{itemize}

Every real series is truncated to a maximum length of 32k time steps.
For longer series, sliding windows are applied. The first 32k steps form the first sample, and subsequent windows are constructed by extending backwards from the tail of the remaining portion.
This yields approximately 43.7M real training sequences.

Synthetic data are generated to cover diverse patterns observed in real time series while keeping sequence lengths within the pre-training window.
Three types are produced, totaling roughly 43M sequences:
\begin{enumerate}
    \item \textbf{Chronos-style synthesis (15M sequences)}: following the protocol of Chronos~\cite{ansari2025chronos}, 4096-step sequences are drawn from Gaussian processes with randomly sampled kernel parameters, introducing multi-scale fluctuation patterns.
    \item \textbf{Long-range GP-interpolated synthesis (15M sequences)}: direct GP sampling at length 32k is computationally prohibitive due to the $\mathcal{O}(n^{3})$ time complexity of dense Gaussian process sampling (cubic in the number of conditioning points).
    We therefore first sample coarse 4096-step sequences and then interpolate groups of eight windows to 32k steps using GP-based probabilistic infilling, which preserves fine-grained stochastic fluctuations rather than performing naive linear interpolation.
    \item \textbf{Non-uniformly sampled synthesis (13M sequences)}: real-world sensor data often contain random gaps caused by device failure or transmission interruptions.
    To mimic this, we generate 4096-step GP sequences, interpolate groups of four windows to approximately 16k steps via GP infilling, and then randomly drop roughly 10\% of the time points.
    The dropped points are not recorded as missing values with a placeholder; they are simply absent from the sequence, requiring the model to infer full dynamics from incomplete observations.
\end{enumerate}

Together, the three synthetic families cover uniform high-frequency signals, long-range continuous patterns, and non-uniformly missing data, complementing those in real data.

\subsection{Contiguous Patch Masking}\label{sec:cpm}

Following TiRex~\cite{auer2026tirex}, we adopt Contiguous Patch Masking (CPM) as the pre-training task, so that the model is exposed to the NAF inference mode already during training.

For each training sample, we uniformly sample a contiguous patch count $c_{\text{mask}} \in \{1, \ldots, c_{\text{mask}}^{\max}\}$ and a masking probability $p_{\text{mask}} \sim U(0, p_{\text{mask}}^{\max})$, partition the $T$ patches of the sequence into $\lceil T / c_{\text{mask}} \rceil$ groups of $c_{\text{mask}}$ patches each, draw an independent Bernoulli variable with parameter $p_{\text{mask}}$ for each group, and extend it to all patches and time steps within that group. Adjacent groups may be masked simultaneously, so the effective masked span can exceed $c_{\text{mask}}^{\max}$.

Masked patches are replaced by placeholders in the input, and their ground-truth values serve only as supervision in the loss; unmasked positions also contribute to the loss. This design exposes the model to an input form in which the prediction region is filled with placeholders, so that it learns to reconstruct a contiguous future from arbitrary historical positions rather than extrapolating only at the sequence terminus, aligning training with the input distribution at NAF inference.

\subsection{Optimizer and Training Schedule}

All three architectures use identical optimization settings.
The learning rate follows a two-phase schedule: it warms up linearly from zero to $1 \times 10^{-3}$ over the first 8k steps, then decays via a cosine schedule to $3.3 \times 10^{-6}$ over the subsequent 212k steps.
The total of 220k steps corresponds to exactly one pass over the full training corpus, so each sample is visited only once. This schedule limits repeated exposure to the same samples, but does not by itself guarantee the absence of overfitting.
We use the Muon optimizer~\cite{liu2025muon} with weight decay 0.1 and gradient clipping norm 1.0.
We do not apply dropout during training.
Complementary to CPM, next-patch prediction ensures that every patch directly contributes to the loss, so that training signals are fully utilized.

HART-36M, pure RWKV-36M, and pure Transformer-35M are pre-trained under exactly the same data, training strategy, and random seed.
This shared setup supports comparison of the architectures under the reported training configuration.
All models are trained on two A800-80G GPUs using Keras 3 and JAX.

\section{Detailed Computational Formulae for Each Block}\label{sec:module-details}

Figure~\ref{fig:architecture} illustrates the internal structure of the two building blocks that compose the HART backbone. This appendix provides the complete computational formulae for each sub-module shown therein. HART introduces minimal modifications to the established architectures of both RWKV and Attention.

\paragraph{Patch Embedding Layer}
\label{sec:patch_details}

The input time series first undergoes $z$-score normalization~\cite{kim2021reversible} to unify sequences of different scales into a comparable numerical range:

\begin{equation}
\tilde{y}_{1:C} = \frac{y_{1:C} - \bar{y}_{1:C}}{\sigma_{y_{1:C}}}
\end{equation}

where $\bar{y}$ and $\sigma_y$ are the mean and standard deviation of the historical series, respectively; these two statistics are retained during inference for inverse normalization in the output head.

The normalized series is segmented into patches of length $P$, yielding a patch sequence $\mathbf{X} \in \mathbb{R}^{B \times T \times P}$.
For each position, we construct an observation mask $\mathbf{O} \in \{0,1\}^{B \times T \times P}$ indicating whether the value is observed, and a missingness mask $\mathbf{N} \in \{0,1\}^{B \times T \times P}$ indicating whether the position is a placeholder to be predicted.
The three tensors are concatenated along the last dimension and then projected to dimension $D$ through a residual MLP.
We follow the skip embedding design of Moirai~\cite{liu2025moirai}:

\begin{equation}\label{eq:patch-embed}
\mathbf{Z} = \mathrm{SiLU}(\mathbf{X}_{\text{cat}} W_{\text{in}}) W_{\text{out}} + \mathbf{X}_{\text{cat}} W_{\text{skip}}
\end{equation}

where $\mathbf{X}_{\text{cat}} = \mathrm{concat}(\mathbf{X}, \mathbf{O}, \mathbf{N}) \in \mathbb{R}^{B \times T \times 3P}$, $W_{\text{in}} \in \mathbb{R}^{3P \times 4D}$, $W_{\text{out}} \in \mathbb{R}^{4D \times D}$, and $W_{\text{skip}} \in \mathbb{R}^{3P \times D}$.
$T = T_{hist} + T_{pred}$ is the total number of patches, where $T_{hist} = \lceil C/P \rceil$ is the number of historical patches and $T_{pred} = \lceil H/P \rceil$ is the number of prediction placeholder patches.
The original inputs in the prediction region are processed as described in Section~\ref{sec:method-pid}; their embeddings incorporate the masks and are not constrained to be zero, while the prediction region contains no observed future values.
The patch embedding layer shares weights between historical patches and prediction placeholder patches.

\subsection{Attention Block}\label{sec:att-block}

Figure~\ref{fig:architecture}(b) depicts the Attention Block, which comprises RMSNorm, gated self-attention, and a SwiGLU feed-forward network with residual connections. It applies sigmoid gating to the self-attention output, following the gated-attention formulation of~\cite{qiu2026gated}.

\subsubsection{Gated Attention}

Given the input $x$, the query, key, value, and gate are obtained via linear projections:

\begin{align}
Q = x W_Q, \quad K = x W_K, \quad V = x W_V, \quad \mathbf{G} = x W_{G}
\end{align}

All projection matrices are $\in \mathbb{R}^{D \times D}$. After applying RoPE positional encoding~\cite{su2024roformer} to $Q$ and $K$, causal self-attention is computed and gated. $\mathbf{C}$ denotes the causal mask:

\begin{align}
A &= \mathrm{softmax}\left(\frac{Q K^\top}{\sqrt{d_h}} + \mathbf{C}\right)V \\
O &= A \odot \mathrm{sigmoid}(\mathbf{G}) \\
O_{\text{out}} &= O W_O, \quad W_O \in \mathbb{R}^{D \times D}
\end{align}

\subsubsection{Feed-Forward Network}

The FFN adopts SwiGLU activation. The only difference from the RWKV-FFN is the absence of token shift; all other aspects remain consistent:

\begin{align}
\mathrm{FFN}(x) &= \bigl(\mathrm{Swish}(x W_{\text{gate}}) \odot x W_{\text{key}}\bigr) W_{\text{out}}
\end{align}

where $W_{\text{gate}}, W_{\text{key}} \in \mathbb{R}^{D \times 4D}$ and $W_{\text{out}} \in \mathbb{R}^{4D \times D}$.

\subsection{RWKV Block}\label{sec:rwkv-block}

Figure~\ref{fig:architecture}(c) depicts the RWKV Block, which consists of LayerNorm, RWKV-TimeMIX (generalized delta rule), and a SwiGLU channel mixer with residual connections. The RWKV Block inherits the full design of RWKV-7~\cite{peng2025rwkv}, with the only modification being the replacement of the activation function in RWKV-FFN from ReLU$^2$ to SwiGLU. Following the standard RWKV-7 architecture, the first RWKV block in the backbone prepends an extra LayerNorm before the TimeMix sub-block for input preprocessing.

\subsubsection{Token Shift}

Both TimeMix and RWKV-FFN employ the token shift mechanism, which takes place after layer normalization and before the respective submodules. Token shift linearly interpolates the current token with the previous token:

\begin{equation}
\mathrm{lerp}(a, b, x) = a + (b - a) \odot x
\end{equation}

The shifted inputs for TimeMix are $x^\square_t = \mathrm{lerp}(x_t, x_{t-1}, \mu_\square)$ for $\square \in \{r, k, v, d, a, g\}$, where each $\mu_\square \in \mathbb{R}^D$ is a learnable mixing factor. RWKV-FFN follows the same pattern, using independent mixing factors $\mu'_k, \mu'_g \in \mathbb{R}^D$.

\subsubsection{LoRA Parameterization in TimeMix}

The TimeMix quantities are computed from the shifted inputs using learned projections, LoRA adapters, and nonlinear transformations:

\begin{align}
a_t &= \mathrm{sigmoid}(\mathrm{LoRA}_a(x^a_t)) \\
\nu_t &= \mathrm{sigmoid}(\mathrm{LoRA}_\nu(x^v_t)) \\
d_t &= \mathrm{LoRA}_d(\tanh, x^d_t) \\
w_t &= \exp(-e^{-0.5} \mathrm{sigmoid}(d_t)) \\
g_t &= \mathrm{LoRA}_g(\mathrm{sigmoid}, x^g_t) \\
k_t &= x^k_t W_k \\
v_{t,l} &= x^v_t W_v \\
r_t &= x^r_t W_r
\end{align}

where $W_k, W_v, W_r \in \mathbb{R}^{D \times D}$. Before head splitting, the removal and replacement keys are $\kappa_t = k_t \odot \xi$ and $\tilde{k}_t = k_t \odot \mathrm{lerp}(1, a_t, \alpha)$, with trainable vectors $\xi, \alpha \in \mathbb{R}^{D}$.
Following RWKV-7, value residual learning interpolates the final value as $v_t = v_{t,0}$ for $l = 0$ and $v_t = \mathrm{lerp}(v_{t,0}, v_{t,l}, \nu_t)$ for $l \ge 1$.

\subsubsection{GDR State Update and Attention Readout}

We split $\{r_t, w_t, \tilde{k}_t, v_t, \kappa_t, a_t\}$ into $h$ heads, each of dimension $d_h = D/h$, and normalize the removal key within each head as $\hat{\kappa}_t = \kappa_t / \lVert\kappa_t\rVert_2$.
All subsequent state-update and readout operations are per-head, with the head index omitted.
The state follows Equation~\ref{eq:gdr}. In a generic kernel using a plus-sign rank-one correction, the corresponding effective vectors are $a_t^{\mathrm{ker}}=-\hat{\kappa}_t$, $b_t^{\mathrm{ker}}=a_t\odot\hat{\kappa}_t$, and $k_t^{\mathrm{ker}}=\tilde{k}_t$; $b_t^{\mathrm{ker}}$ is therefore not an independently projected vector.
The attention result $p_t$ is read from the state via the receptance vector:

\begin{align}
u_t &= \bigl(r_t \cdot (\rho \odot \tilde{k}_t)^\top\bigr) v_t, \quad \text{bonus term} \\
p_t &= \mathrm{LayerNorm}(r_t S_t^\top) + u_t
\end{align}

where $\rho \in \mathbb{R}^{d_h}$ is the bonus weight. After concatenating all heads, the final output of TimeMix is obtained by gating and output projection:

\begin{equation}
o_t = (g_t \odot p_t) W_o, \quad W_o \in \mathbb{R}^{D \times D}
\end{equation}

\subsubsection{RWKV-FFN (SwiGLU Version)}

RWKV-FFN employs token shift and SwiGLU activation. The only difference from the original RWKV-7 is the replacement of ReLU$^2$ with SwiGLU:

\begin{align}
k'_t &= \mathrm{lerp}(x'_t, x'_{t-1}, \mu'_k) W_{k'} \in \mathbb{R}^{4D} \\
g'_t &= \mathrm{lerp}(x'_t, x'_{t-1}, \mu'_g) W_{g'} \in \mathbb{R}^{4D} \\
o'_t &= \bigl(\mathrm{Swish}(g'_t) \odot k'_t\bigr) W_{v'} \in \mathbb{R}^{D}
\end{align}

where $\mathrm{Swish}(x) = x \cdot \mathrm{sigmoid}(x)$, and $W_{k'}, W_{g'} \in \mathbb{R}^{D \times 4D}$, $W_{v'} \in \mathbb{R}^{4D \times D}$.

\subsection{Summary of Architectural Differences from the Originals}\label{sec:detail-block}

HART modifies existing modules only in the following respects:

\begin{enumerate}
    \item \textbf{RWKV Block}: Only the activation function in RWKV-FFN is changed from ReLU$^2$ to SwiGLU to maintain a consistent activation function choice with the Attention Block. The entire TimeMix (including GDR, token shift, LoRA parameterization, value residual learning, and attention readout) retains the original design of RWKV-7.
    \item \textbf{Attention Block}: We use gated self-attention and a SwiGLU FFN, with the formulations specified above.
\end{enumerate}

Based on these components, we assemble them into our proposed HART architecture. 

\end{document}